\documentclass[10pt,letterpaper]{article}

\usepackage[nonatbib,preprint]{neurips_2024}
\usepackage[sort,compress,square,numbers]{natbib}

\usepackage{custom}
\usepackage{glossaries}
\glsdisablehyper
\newacronym{pca}{PCA}{Principal Component Analysis}
\newacronym{dft}{DFT}{Discrete Fourier Transform}
\newacronym{aee}{AEE}{Average Endpoint Error}

\graphicspath{{../imgs/}{../}}

\usepackage[compact,tiny]{titlesec}
\makeatletter
\DeclareMathSizes{\@xpt}{9}{7}{5}
\makeatother

\title{
Many Perception Tasks are Highly Redundant Functions of their Input Data
}

\author{%
Rahul Ramesh$^{*}$
\And
Anthony Bisulco$^{*}$
\And
Ronald W. DiTullio
\And
Linran Wei
\AND
Vijay Balasubramanian
\And
Kostas Daniilidis
\And
Pratik Chaudhari \\
\AND 
\textnormal{Email: \{rahulram, abisulco, rdit, linran, vijay, kostas, pratikac\}@upenn.edu}
}

\definecolor{warningcolor}{RGB}{204,51,20}
\graphicspath{{../}}

\begin{document}

{\footnotesize \listoftodos}

\clearpage
\maketitle

{ \let\thefootnote\relax\footnotetext{$^{*}$Equal Contribution}}

\begin{abstract}
We show that many perception tasks, from visual recognition, semantic segmentation,  optical flow, depth estimation to vocalization discrimination, are highly redundant functions of their input data.
Images or spectrograms, projected into different subspaces, formed by orthogonal bases in pixel, Fourier or wavelet domains, can be used to solve these tasks remarkably well regardless of whether it is the top subspace where data varies the most, some intermediate subspace with moderate variability---or the bottom subspace where data varies the least.
This phenomenon occurs because different subspaces have a large degree of redundant information relevant to the task.
\end{abstract}


\section{Introduction}
\label{s:intro}
Suppose we are given a dataset with inputs that are vectors in Euclidean space and outputs denoting each input's ground-truth category.
The textbook procedure for modeling this data often begins with principal component analysis (PCA~\citep{hotelling1933analysis}).
PCA projects inputs onto the ``principal'' subspace where the variance of the projection is maximal.
PCA can be used to reduce the dimensionality, remove the effects of noise, and, as every reader has done in the past, fit a model for predicting the labels using these salient features.
The larger the explained variance of PCA, the smaller (hopefully) the information about the labels thrown away by the projection.
The smaller the dimension of the principal subspace, the more robust the model is to data variations in the discarded subspace.
This is why any data science textbook teaches its readers to identify the ``elbow'' in a scree plot as the size of the principal subspace~\citep{james2013introduction}.

This paper shows that for many perception tasks—from visual recognition, semantic segmentation, optical flow, depth estimation to auditory discrimination—one can accurately predict the output using non-salient features. The principal subspace is most predictive of these tasks. However, the predictive ability of any other subspace is remarkably high. These perception tasks are, therefore, highly redundant functions of their input data.

We examine this phenomenon through different lenses, using ideas from signal processing, information theory, and neuroscience. \cref{s:results} discusses our results, where we identify common themes and important differences across these modalities. \cref{s:discussion} discusses how these observations are related to existing results in the literature and their implications. Finally, \cref{s:methods} details the tools used in our analysis and the different tasks.


\section{Results}
\label{s:results}

We next describe our findings using a broad range of evidence, analysis, and discussion.
We use the CIFAR-10~\citep{krizhevsky2009learning} and ImageNet~\citep{dengImagenetLargescaleHierarchical2009} datasets for classification;
the Cityscapes dataset~\cite{cityscapes} for semantic segmentation and  depth estimation;
the ADE20K dataset for semantic segmentation~\cite{ade20k};
and an augmented version of the M3ED dataset~\cite{m3ed} for optical flow, depth estimation and semantic segmentation. These tasks are diverse and complex, e.g.,
M3ED contains data from many natural scenes, including cars driving in urban, forest, and daytime/nighttime conditions.
\cref{s:setup} elaborates upon our experimental setup.

We study these tasks using the PCA, Fourier and wavelet bases and seek to understand which basis elements are important for these tasks.
We create different indices that order the elements of these bases.
For PCA, we sort by explained variance, lower indices have high explained variance. For the Fourier basis, lower indices correspond to smaller radial frequencies.
For the wavelet basis, lower indices correspond to the smallest scales. 
See~\Cref{s:app:index} for more details on how the indices are constructed.%

\subsection{Both input data and the task are effectively low-dimensional. Elements of the Fourier and wavelet bases have a large overlap with those of PCA.}
\label{s:result_1}

\begin{figure}[!htpb]
    \centering
    \includegraphics[width=0.95\linewidth]{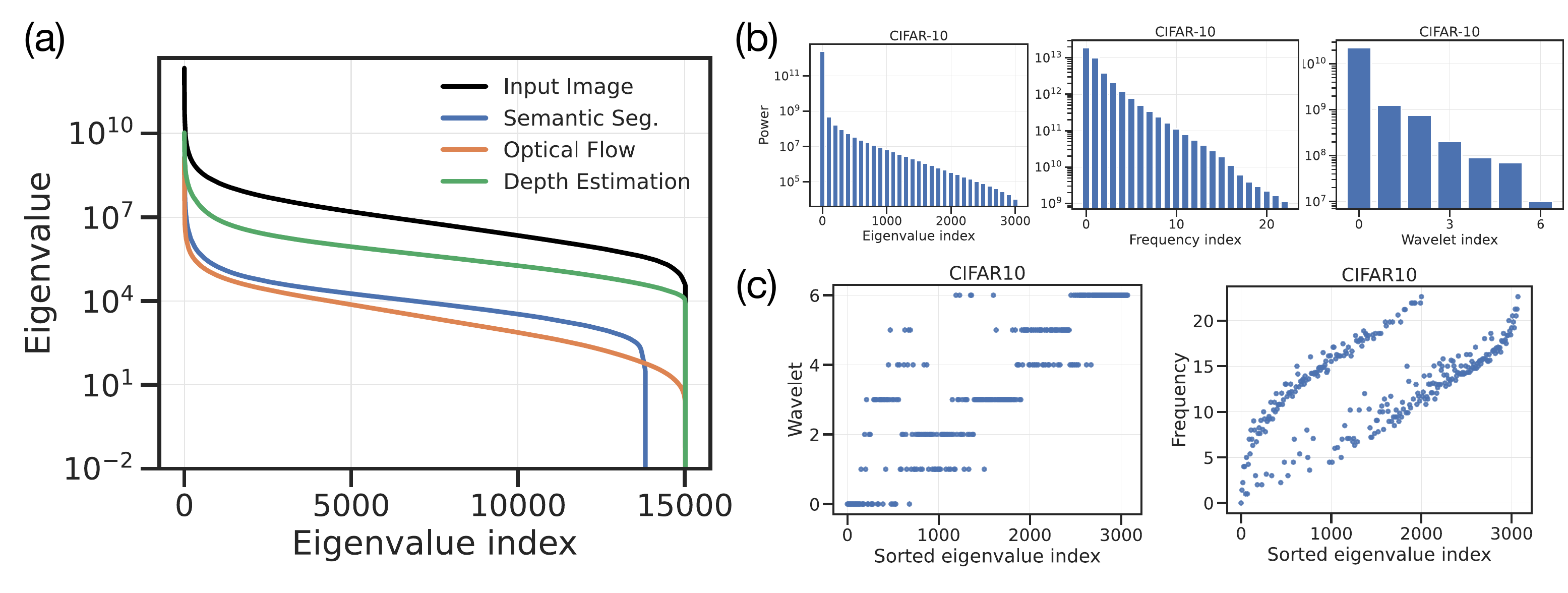}
    \caption{
    \textbf{(a)} Eigenvalues of the pixel-wise covariance matrix for inputs and outputs of different tasks are spread across a large range and decay quickly.
    \textbf{(b)} Variance or energy decays quickly with an increase in the index for PCA, Fourier and wavelet bases.
    \textbf{(c)} Index of wavelet or Fourier basis element (y-axis) that has the highest amplitude for images projected onto a PCA eigenvector of a particular index (x-axis). High Fourier and wavelet indices (large radial frequency and large scale, respectively) correspond to PCA eigenvectors with higher indices (or smaller eigenvalues).
    }
    \label{fig:result_1}
\end{figure}

\begin{wrapfigure}[11]{r}{0.4\linewidth}
    \centering
    \includegraphics[width=\linewidth]{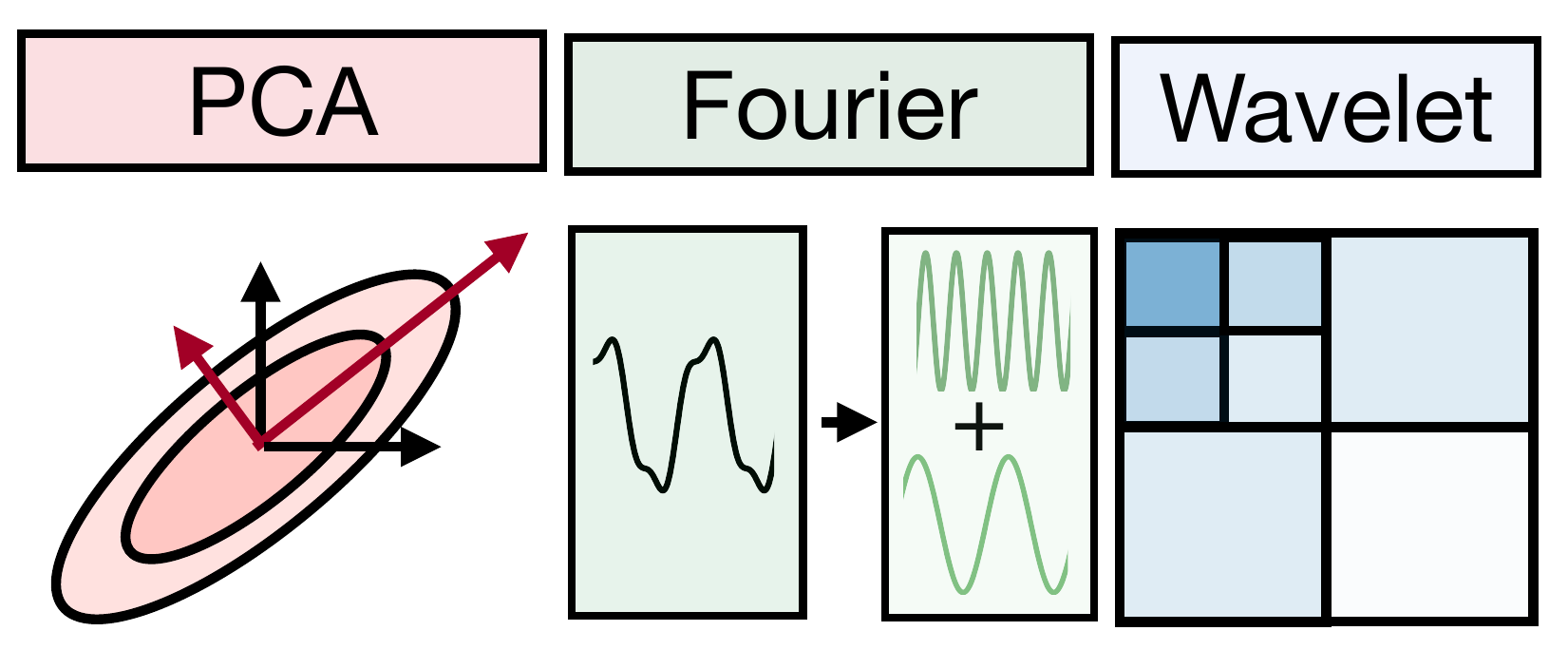}
    \caption{Schematic of Principal Components Analysis, Fourier and wavelet basis.}
\end{wrapfigure}
\cref{fig:result_1} (a) shows that for all datasets and modalities, inputs measured in different bases, namely pixel, Fourier, and wavelet space, exhibit a characteristic low-dimensionality.
Eigenvalues are spread across a very large range---more than $10^6$---for many of these tasks.
When eigenvalues are sorted by their magnitude, there is a sharp ``elbow'', i.e., very few dimensions are necessary to capture most of the variance in the data, e.g., for CIFAR-10 which consists of 32$\times$32 RGB images, the subspace corresponding to the first 5 eigenvalues of the pixel-wise covariance matrix has 90\% explained variance and the remaining 3067 dimensions capture a mere 10\% of the variance.
The eigenspectra of all these datasets contain a long tail of eigenvalues that are spread linearly on a logarithmic scale.
The numerical difficulty of any optimization algorithm, even linear regression, is governed by the condition number of the input covariance matrix~\cite{boyd2004convex}. The optimization problem underlying perception tasks is, therefore, ill-conditioned.
This can result in long training times and under-fitting to the signal in the tail.

For some tasks, namely optical flow, depth estimation, and semantic segmentation, the ground-truth output can also be considered an image to calculate its eigenspectrum.
Again, the characteristic low-dimensional pattern is evident in all bases.
It is well known that amplitude spectra of natural images decay as $\sim 1/\abs{f}$ with the frequency $f$; see~\cref{s:discussion}. We see a similar phenomenon for dense output tasks.
The new observation here is that the ground-truth labels of many tasks are also effectively low-dimensional.

PCA, Fourier, and wavelet are three different orthogonal bases. \cref{fig:result_1} (b) shows that there is a large degree of overlap between inputs and outputs projected onto different subspaces in these bases, i.e., coefficients corresponding to the principal subspace of PCA are highly correlated with those of the Fourier basis at small radial frequencies or wavelets of small scales.
Even if the three bases are linear transformations of the original pixel space, we would not have expected them to be aligned like this.
Certainly, PCA is a dataset-dependent basis, while Fourier and wavelet bases are universal; we discuss this in~\cref{s:PCA}.

\subsection{The principal subspace of the input data is most predictive of the task. However, the subspace with the least explained variance is also remarkably predictive. Even a random subspace is predictive of the task.}
\label{s:result_2}
\begin{figure}[!htb]
\centering
\includegraphics[width=0.95\linewidth]{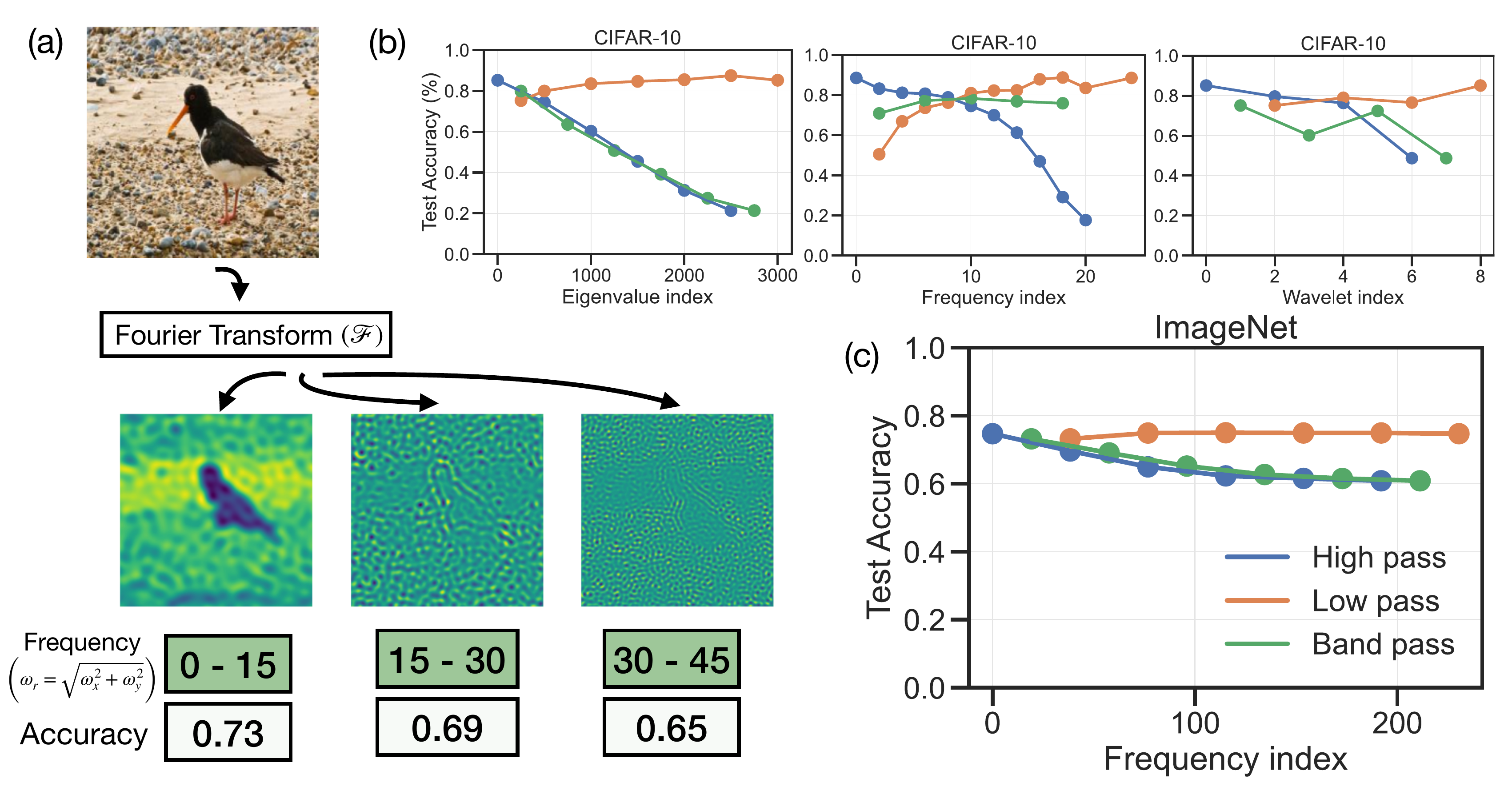}
\includegraphics[width=0.95\linewidth]{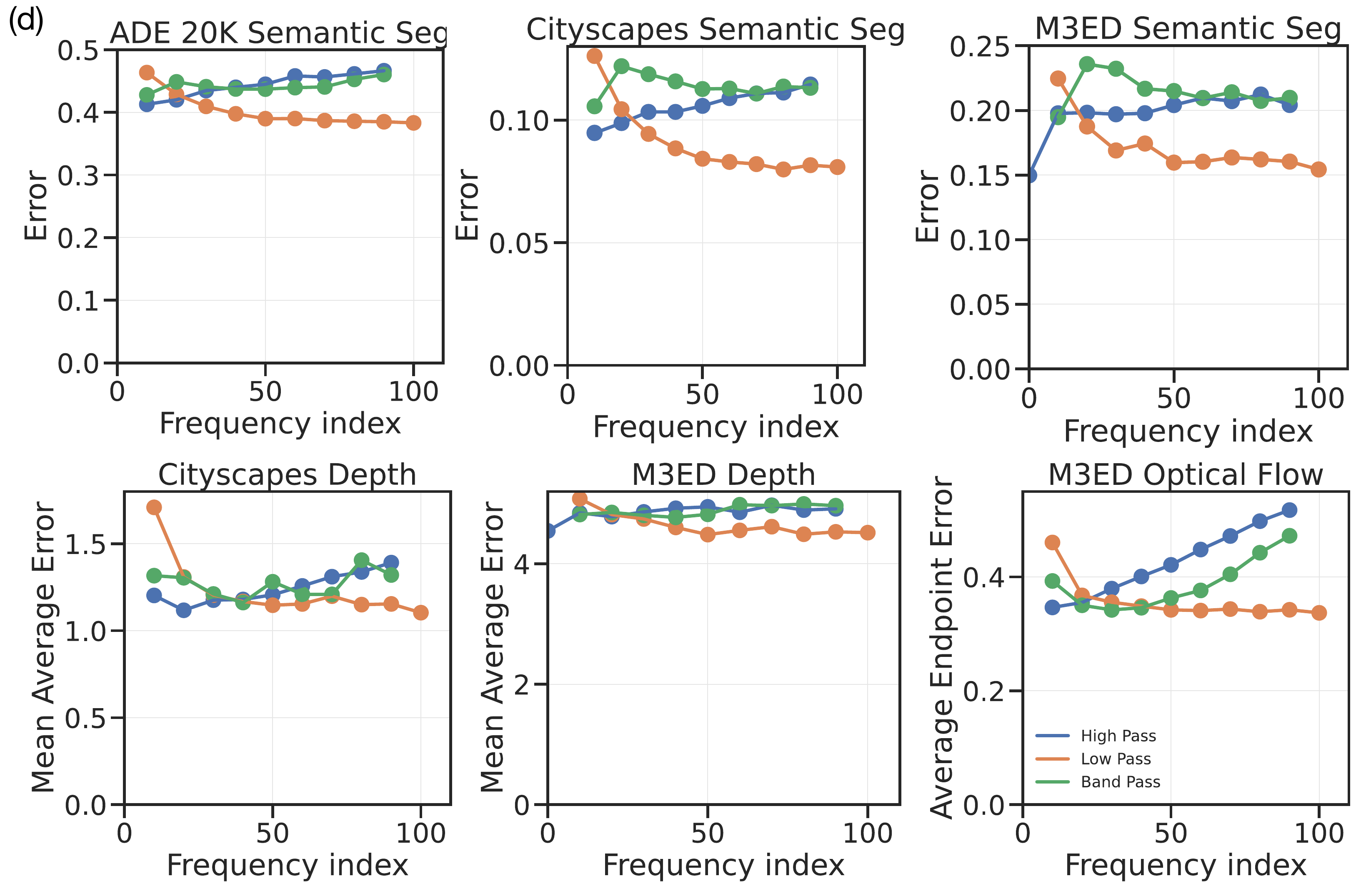}
\caption{
Panel \textbf{(a)} shows that the image, when projected on a high frequency band (30--45) cannot be recognized by the human eye; and yet a network trained on such images can get more than 65\% test accuracy.
We show the test accuracy (for CIFAR10 \textbf{(b)} and ImageNet \textbf{(c)}) of networks trained on images projected onto different subspaces.
Remarkably, for ImageNet, all frequency bands achieve more than 60\% accuracy.
Almost all PCA subspaces, radial frequencies and scales are useful for image classification on CIFAR-10 and ImageNet; observe that low pass, band pass and low-index high pass regimes all obtain good test accuracy.
However, the head of the spectrum usually contains more discriminative information than the tail.
\textbf{(d)} For dense perception tasks such as semantic segmentation, optical flow and depth prediction, the results are consistent with classification, i.e., the information for the task is also present redundantly across the spectrum.
Many frequency bands result in remarkably low errors on these tasks. Error barely improves with index for low pass filters, indicating diminishing returns on these tasks as higher frequencies are included in the data.}
\label{fig:result_2}
\end{figure}
We trained deep networks on inputs projected onto different subspaces. We create these subspaces using ``bands'' of basis elements formed by the PCA, Fourier, and wavelet bases.
Each band contains a contiguous block of basis elements sorted by their index.
A low-band pass with an index $i$ consists of a subspace that contains all eigenvectors/frequencies/scales with an index smaller than $i$.
The principal subspace where the explained variance/energy per dimension is the highest (see~\cref{s:result_1} (b)) lies in a low-band pass with a small index $i$.
These bands are cumulative, so the principal subspace also lies in a low-band pass with a large index $i$.
The high-band pass at index $i$ is a subspace that consists of all eigenvectors/frequencies/scales with an index bigger than $i$.
A high-band pass with a large index $i$ contains the subspace with the smallest explained variance per dimension; a high-band pass with a small index $i$ also contains this subspace.
A band pass of index $i$ corresponds to a subspace formed by a few sorted eigenvectors/frequencies/scales to the left and right of $i$ with width specified for each experiment.
An $i$-$j$ band pass corresponds to a subspace formed by eigenvectors/frequencies/scales with an index between $i$ and $j$.
The subspace that corresponds to index $i$ does not have any overlap with that of index $(i+1)$%
\footnote{A Butterworth~\cite{Butterworth1930} filter of order 5 is used to avoid ringing effects produced from a box filter; the cutoff frequency is set at the 3 dB point. Due to this, there is a small overlap in our bands for the Fourier basis; there is no such overlap in the PCA or wavelet bands.}.

\begin{figure}[!tb]
    \centering
    \includegraphics[width=0.95\linewidth]{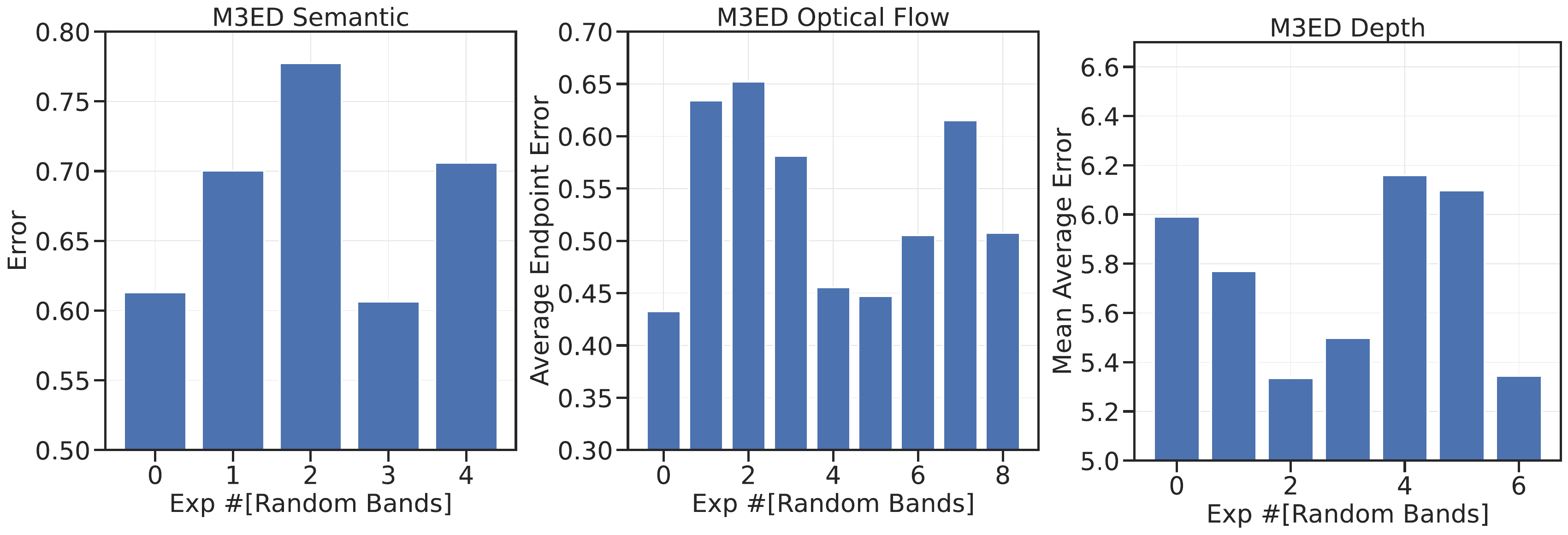}
    \caption{Error on perception tasks is remarkably low even when networks are trained on images projected onto a random linear subspace; see \ref{app:rnd_bands} for the parameters of the randomly chosen center frequencies and widths for the band pass filter.
    This suggests that information for the task is present throughout the spectrum. The ``explained'' power is less than 20\% for all bands of randomly chosen frequencies, for all three panels.
    }
    \label{fig:random_pass_band}
\end{figure}

\paragraph{Classification tasks.}
\cref{fig:result_2} (a) and (c, green) show that different bands of the Fourier basis have remarkably high accuracy on ImageNet. In~\cref{fig:result_2} (a), the original image is unrecognizable to the human eye when projected into higher frequency bands. But a deep network can be trained on such images, and it gets 65\% test accuracy in the 30--45 radial frequency band pass.
\cref{fig:result_2} (b, green) shows that the test accuracy on CIFAR-10 is largely the same for different bands for the Fourier (middle) and wavelet (right) bases, but it drops for PCA (left).%
\footnote{This can be understood by looking at~\cref{fig:explained_variance_power}. As the index increases, the number of orthogonal components in the PCA band pass is unchanged, so the explained variance of different bands decreases. But for Fourier and wavelet bases, the number of frequencies/scales increases sharply with the index. This is because high radial frequencies span a large number of spatial frequencies.
}
In short, the predictive ability of a band is correlated with its explained variance/power.
However, many of these bands are non-trivially predictive of the task. For CIFAR-10, the last band pass of PCA has 20 \% accuracy.
\cref{fig:result_2} (b, orange) shows that the test accuracy increases with the low pass index $i$ as the dimensionality of the subspace grows; this trend also holds for ImageNet in~\cref{fig:result_2} (c, orange).
Adding new basis elements, eigenvectors, frequencies, or scales has diminishing returns on accuracy.
This trend is exactly reversed for high pass---accuracy drops with the index $i$ as the explained variance/power of these subspaces decreases.
\cref{fig:random_pass_band} shows that band passes created from a random subset of the basis are also predictive of the task.

Therefore, the principal subspace is usually the most predictive (green curves decrease with index, except for the Fourier basis).
The tail of the spectrum also has remarkable predictive ability.
Essentially, any subspace of the input data is predictive of these tasks.

\paragraph{The above trends also hold for optical flow estimation, depth estimation and semantic segmentation.}
See~\cref{fig:result_2} (d).\footnote{Also see~\cref{fig:perc_extra} for experiments with a different neural architecture and different filters for some of the tasks.}
For optical flow estimation, the Average Endpoint Error (AEE) is smaller for low frequency bands (orange) than high frequency bands (blue).
Error decreases and then saturates with increasing index for low pass filters (orange).
AEE for high pass filters (blue) continues to increase, essentially linearly, with the index.
This might indicate that the low-frequency spectrum contains more predictive features than the high frequency spectrum, where, we suspect, features are redundant for optical flow.
This is consistent with existing results that have argued that low-frequency information is important for motion perception~\cite{Shi2020WithoutLS}.
Trends for semantic segmentation and depth estimation are similar.  Similar to flow, previous literature on depth has shown the importance of low-frequency content for depth estimation ~\cite{kane2014limits}. Error (1-Accuracy) quickly saturates for low pass filters (orange) and decreases roughly linearly for high pass filters (green).
Error of different bands (blue) increases slightly with frequency.

\paragraph{The signal is stronger than the noise, even in the tail.}
\begin{wrapfigure}[15]{r}{0.5\linewidth}
    \centering
    \vspace*{-1em}
    \includegraphics[width=0.75\linewidth]{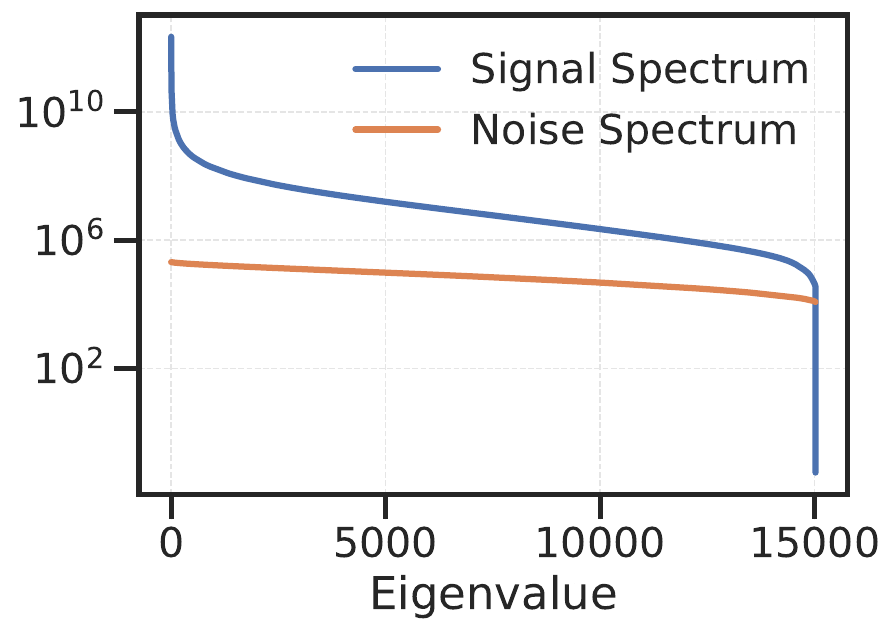}
    \caption{The amplitude of noise in images from the M3ED dataset is roughly constant across the spectrum and smaller than the signal, even in the tail. Signal-to-noise ratio (SNR) is larger in the head than in the tail.} 
    \label{fig:image_noise}
\end{wrapfigure}
Statistical wisdom is to avoid using the tail of the spectrum for making predictions due to noise~\cite{noise1,noise2,noise3} either due to low signal coming from unobserved data (rare categories), or due to high noise of the acquisition process (camera).
Surprisingly, the tail of the spectrum is predictive for so many tasks.
We used the method of~\citet{ivhc_noise}, which estimates the level of additive white Gaussian noise in the image by finding homogeneous patches in the image frame to estimate the variance of intensity there.
For one of our datasets, M3ED, the variance is $\sigma^2=1.41$.
This is shown in~\cref{fig:image_noise} along with the original spectrum for comparison.
As expected, the signal-to-noise ratio (SNR) is larger in the head.
But the noise magnitude is constant over different eigenvalues, and it is well below the magnitude of the smallest eigenvalue.

\begin{figure}[!ht]
    \centering
    \includegraphics[width=0.30\linewidth]{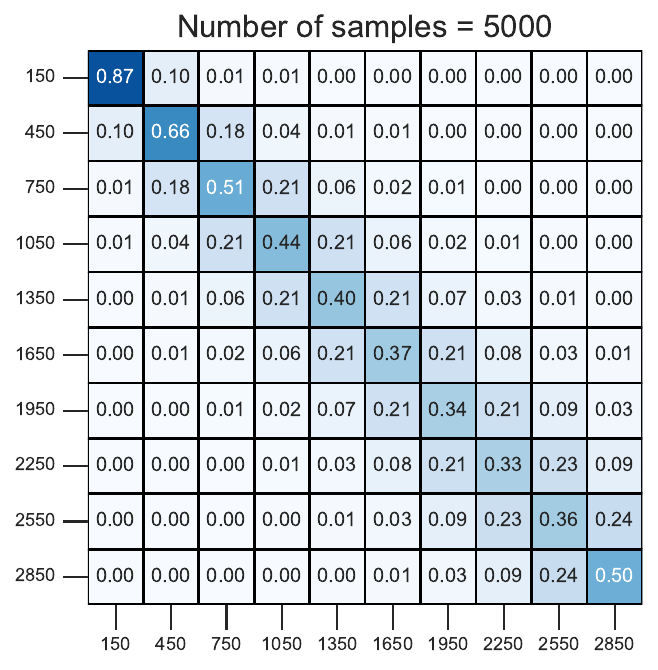}
    \includegraphics[width=0.30\linewidth]{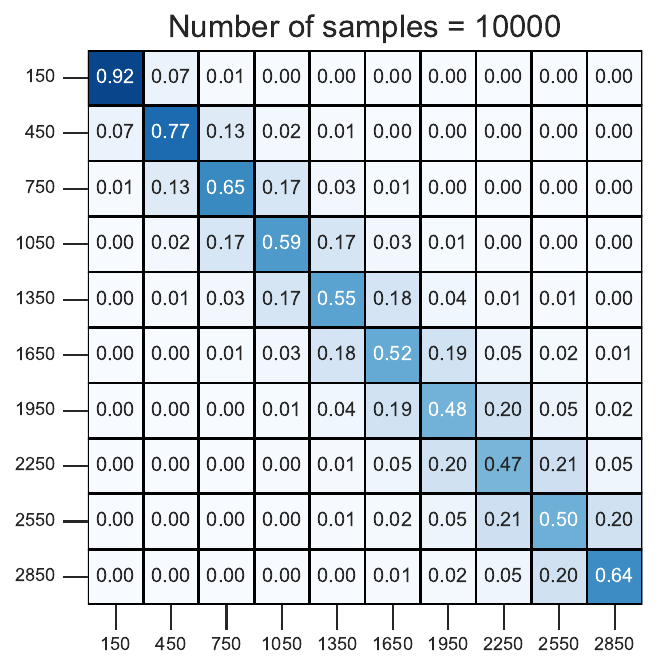}
    \includegraphics[width=0.30\linewidth]{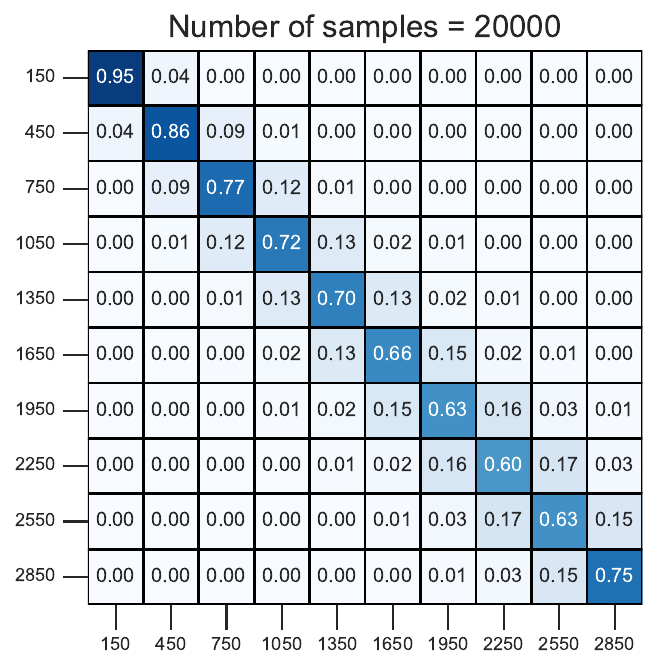}
    \caption{We plot the average similarity between bands of eigenvectors estimated from two different datasets. Both the head and the tail are similar across different draws of the dataset, as indicated by the diagonal entries of the matrix. The head of the spectrum is more stable than the tail. This plot suggests that the entire spectrum is stable and that the signal is stronger than the noise even in the tail.
    }
    \label{fig:stable_tail}
\end{figure}

We make a similar observation from the experiments for  ~\cref{fig:stable_tail}.
We use bootstrap sampling to create multiple datasets with $n=5000,10000,20000$ samples drawn from CIFAR10.
We create 50 datasets for each dataset of size $n$.
For each dataset we compute its eigenvectors, which are divided into 10 bands.
Each band of $d$ eigenvectors defines a $d$-dimensional subspace and the similarity between two bands of eigenvectors $V_1$ and $V_2$ is $\frac{1}{d} ||V_1^T V_2||_F$; the similarity lies between 0 and 1 depending on the extent of overlap between the two subspaces.
In~\cref{fig:stable_tail}, we compute the average similarity between every band of eigenvectors computed using one dataset to that computed using another dataset. The bands of eigenvectors span similar subspaces even if they are computed using different datasets.

\paragraph{Redundancy in vocalization discrimination tasks.}
\citet{bregmanbook} showed that temporal regularity, i.e., patterns in frequency changes over time, plays a critical role in auditory perception.
\citet{ditullio2022time} argued that this could be because of temporal regularities in vocalization; slowest frequencies can be used to discriminate between rhesus macaque vocalizations.

\begin{wrapfigure}[19]{r}{0.6\linewidth}
    \centering
    \vspace*{-0.4em}
    \includegraphics[width=0.75\linewidth]{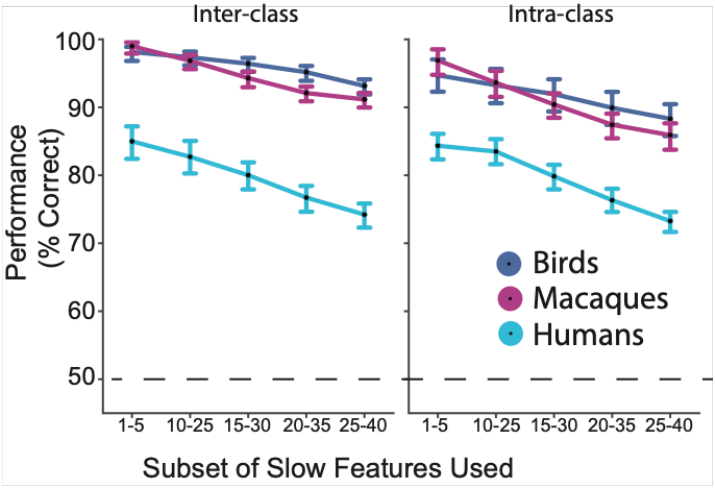}
    \caption{{Vocalizations can be discriminated well, using even fast SFA features}.
    Accuracy across 200 pairs of vocalizations (error bars are 95\% confidence intervals). Left figure is for inter-class pairs, i.e., vocalizations from two different human speakers, macaque speakers, or species of bird. Right figure is for intra-class pairs, i.e., different numerical digits from the same human speaker, utterances from the same macaque, or songs from the same species of bird.}
    \label{fig:AudioResult}
\end{wrapfigure}
We are interested in understanding if faster temporal frequencies can also be used for this task.
We studied bird songs of four species (American yellow warbler, blue jay, great blue heron, and house finch) ~\citep{zhao2017automated}, macaque vocalizations~\citep{fukushima2015distributed}, and human vocalizations of MNIST digits~\citep{becker2024audiomnist} using slow feature analysis (SFA)~\citep{wiskottSlowFeatureAnalysis2002}.
SFA features correspond to the eigenvectors corresponding to the smallest eigenvectors in the time derivative of the auditory stimulus.
\cref{fig:AudioResult} shows that a multi-layer perceptron trained using faster features can also discriminate between different vocalizations quite well.
Performance decreases quite slowly as faster features are used (the confidence intervals overlap), and it is well above chance. This observation is consistent with visual perception tasks.

\subsection{Different PCA subspaces have redundant and synergistic information about the classification task with very little unique information.}
\label{s:result_3}

We next use the classification task to see that the observation above, although surprising, may not be due to any specific properties of deep networks. It seems to be inherent to the data.

\begin{figure}[!tb]
    \centering
    \includegraphics[width=0.99\linewidth]{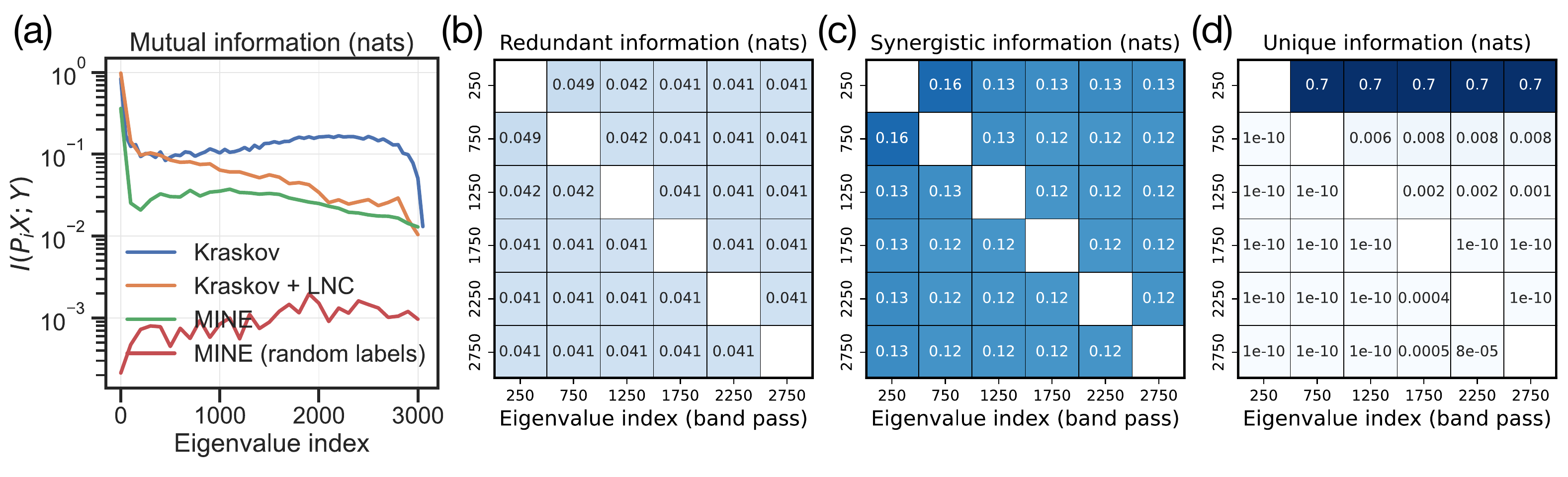}
    \caption{\textbf{(a)} Images projected onto different PCA subspaces (each consisting of 10 contiguous basis elements) have non-trivial mutual information with the labels, across the spectrum.
    Kraskov and Kraskov-LNC are non-parametric estimators of mutual information, MINE uses a neural network\cite{belghazi2018mine}.
    While the principal subspace has the highest mutual information with the labels, the tail also has non-trivial mutual information. For comparison, the mutual information with randomly permuted labels is much smaller.
    Partial information decomposition for CIFAR-10 suggests that different bands have high amounts of redundant \textbf{(b)} and synergistic information \textbf{(c)}. Cell $(i,j)$ corresponds to the PID decomposition for the task's subspaces $i$ and $j$. Unique information is much smaller \textbf{(d)} for anything other than the principal subspace.
    We note that (b,c,d) assume that inputs and labels are jointly Gaussian, which could be a poor approximation.
    Fourier and wavelet basis trends are similar; see~\cref{fig:pid_cifar_2}.
    }
    \label{fig:mi_pid_cifar}
\end{figure}

\paragraph{Labels have high mutual information with every subspace of the input data.}
\begin{wrapfigure}[14]{r}{0.55\linewidth}
    \centering
    \vspace*{-0.5em}
    \includegraphics[width=0.46\linewidth]{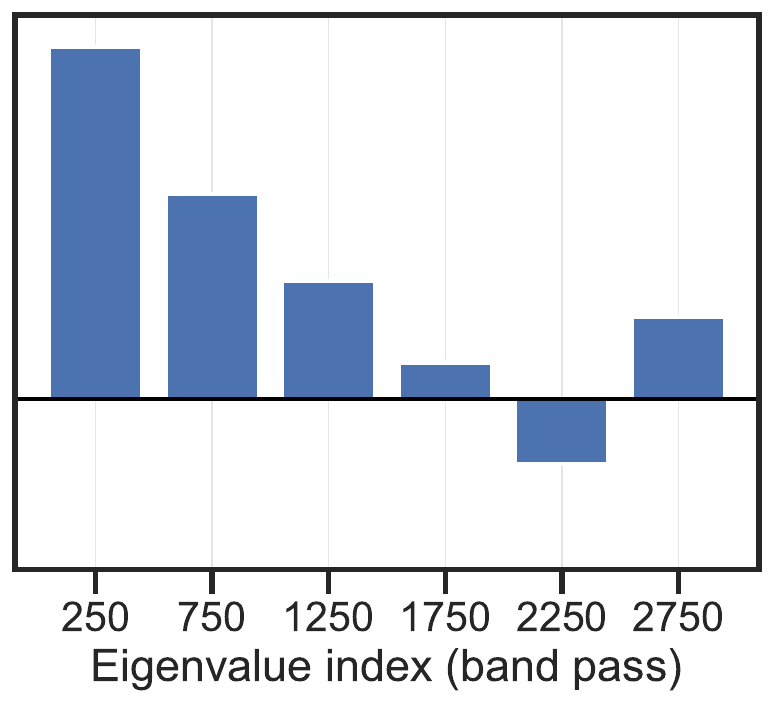}
    \includegraphics[width=0.46\linewidth]{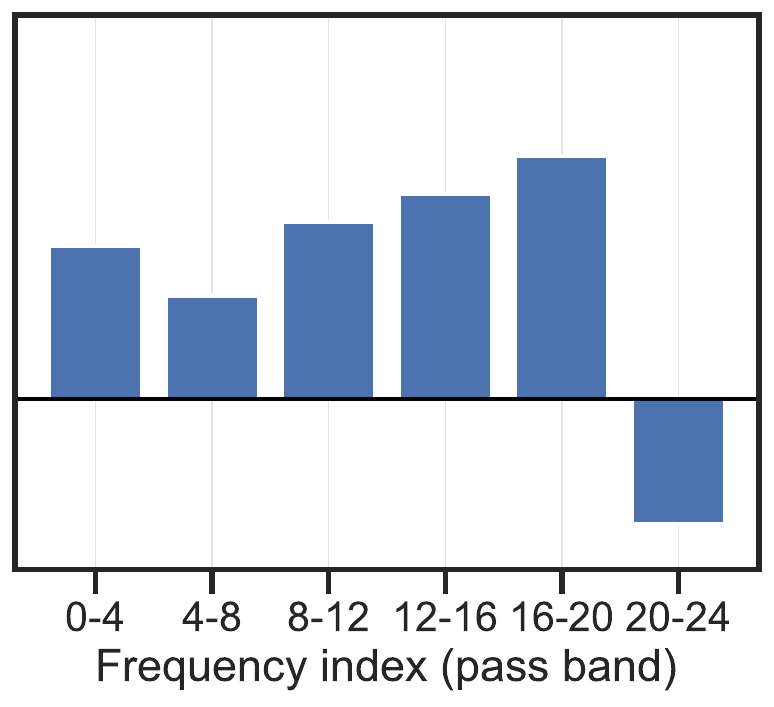}
    \caption{SHAP values of different PCA and frequency bands: The head of the spectrum is the most important, but the tail is also necessary for classification on CIFAR-10. The horizontal black line corresponds to a SHAP value of 0.
    }
    \label{fig:shapley_cifar}
\end{wrapfigure}
\cref{fig:mi_pid_cifar} (a) shows the mutual information $I(Y; P_i X)$ of labels $Y$ with input images $X$ from CIFAR-10, that are projected onto the $i^{\text{th}}$ eigenvector of the pixel-wise input covariance matrix; \cref{s:methods} gives more details.
The Kraskov estimator~\citep{kraskov2004estimating} calculates the mutual information of the discrete labels with the (scalar) projected images, using a $k$-nearest neighbor estimator of the entropy; LNC stands for local non-uniformity correction of the entropy estimator~\citep{gao2015efficient}.
MINE~\citep{belghazi2018mine} is a neural-network based estimator of mutual information.
Both these estimates of $I(Y; P_i X)$ suggest that the principal subspace has high mutual information with the labels. The mutual information is also large for eigenvectors in all the other subspaces. The explained variance of the high-index eigenvectors is very small, so one might expect $I(Y; P_i X)$ to decay strongly for those.
In short, any subspace of the input data should be predictive of the ground-truth labels, \cref{fig:result_2} provided experimental evidence for this.
This finding is corroborated by SHapley Additive exPlanations (SHAP) values in~\cref{fig:shapley_cifar}; also see~\cref{s:methods}.

\paragraph{Partial information decomposition for identifying redundant, unique and synergistic information}
Both the eigenspectrum's head and tail have information pertinent to classification, but we do not know if this is the same kind of information. We used the concept of partial information decomposition (PID~\citep{williams2010nonnegative}) to investigate this. Given variables $X_1$ and $X_2$, PID decomposes the mutual information $I(X_1, X_2, ; Y)$ into redundant ($R$), synergistic ($S$) and unique information ($U_1$ and $U_2$ corresponding to $X_1$ and $X_2$ respectively).
The total mutual information $I(X_1,X_2; Y) = R + S + U_1 + U_2$.%
\footnote{We discuss PID in~\cref{s:methods} further.
Redundancy is the minimum information about $Y$ provided by either variable; it is independent of correlation between the variables.
Synergy $S$ is the extra information contributed by the weaker source when, the stronger
source is known and can either increase or decrease with correlation between sources; typically jointly Gaussian random variables are net synergistic~\citep{barrett2015exploration}.
}\textsuperscript{,}\footnote{It is difficult to calculate PID, or its variants,
using samples~\citep{latham2005synergy}.
We approximated that $X_1, X_2$, and  $Y$ are jointly Gaussian.
Note that the ground-truth label $Y$ is a categorical random-variable.
But we nevertheless plough forward with this approximation.
It is reassuring that this analysis corroborates~\cref{fig:result_2}, so this approximation is not entirely invalid.
}
Different PCA bands correspond to random variables $X_1$ and $X_2$ for us.
The mutual information of each band decomposes as $I(X_1; Y) = R + U_1$. The fact that $I(X_1, Y)$ for different eigenvectors in~\cref{fig:mi_pid_cifar} (a) has a comparable magnitude as that of any cell in~\cref{fig:mi_pid_cifar} (b) suggests that different PCA bands of CIFAR-10 have a lot of redundant information with the labels. This is borne out by~\cref{fig:mi_pid_cifar} (d), which shows that low-index PCA bands (high explained variance) have larger unique information than bands in the tail.
Synergistic information is usually harder to interpret. In this case, it is large for any two PCA bands.
This analysis suggests that the observations in~\cref{fig:result_2} are due to inherent properties of the input data and the ground-truth labels.

\subsection{Despite redundant and synergistic information across the entire spectrum, a deep network predominantly uses information in the head}
\begin{figure}[!th]
    \centering
    \includegraphics[width=0.4\linewidth]{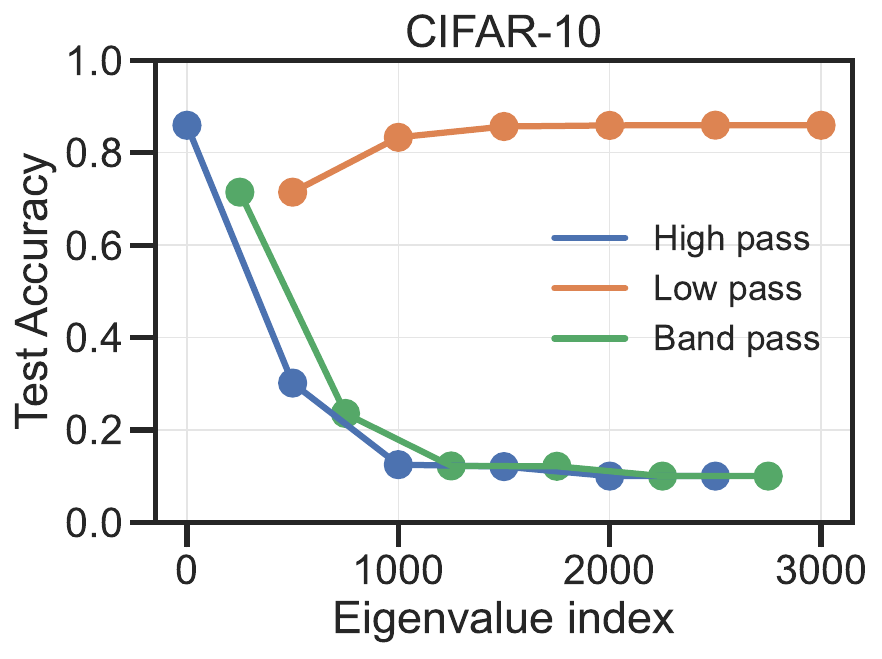}
    \includegraphics[width=0.4\linewidth]{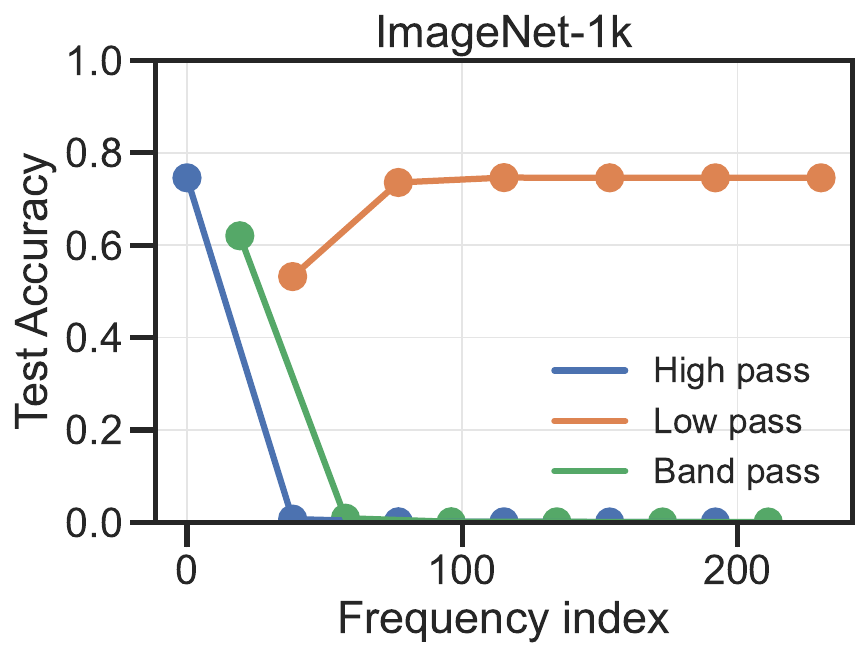}
    \includegraphics[width=0.85\linewidth]{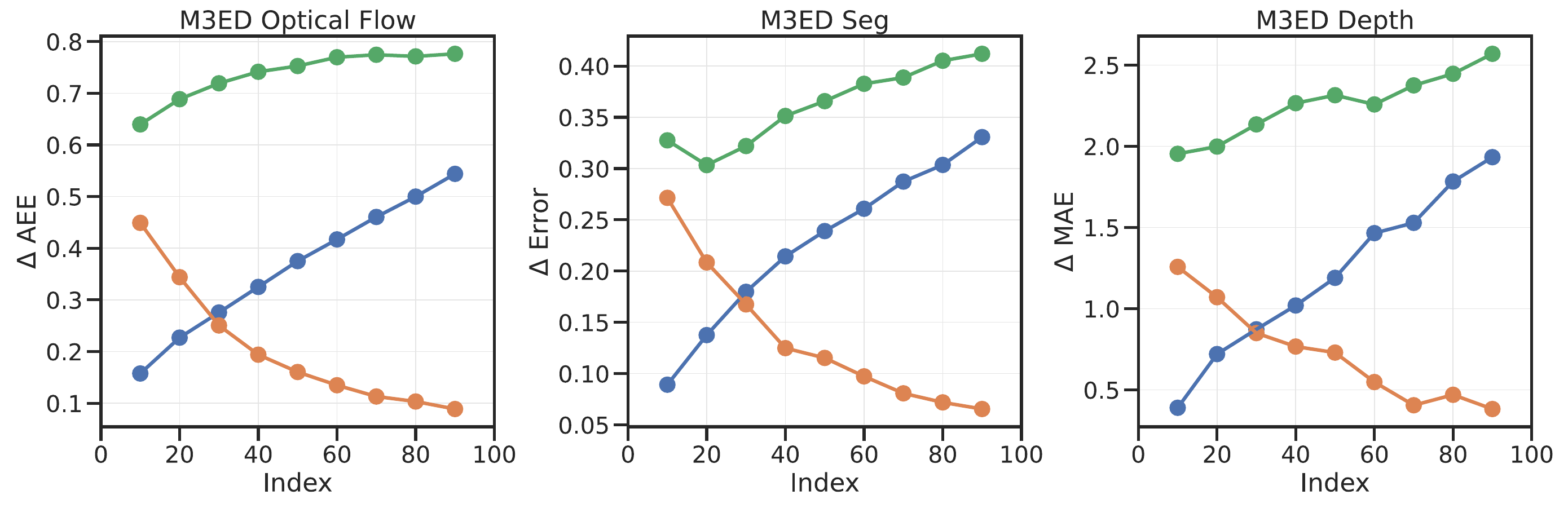}
    \caption{{Trained networks predominantly use information present in the head of the spectrum. They largely ignore the tail}. \textbf{Top}: Accuracy on ImageNet drops to 0.1\% if we exclude radial frequencies smaller than 30 at test time.
    \textbf{Bottom}: 
    For regression tasks, we show the difference between the error of a network trained on every band with a network trained on a particular band; the error for both networks is computed using test images projected onto a particular band.
    Error differential increases with index for band pass and high pass; the network, therefore, predominantly uses information in the head.
    Also, see~\cref{fig:app:feature_space} for a similar result using features from different layers in a trained network.
    }
    \label{fig:result_4}
\end{figure}

The performance of low pass bands improves with the index in~\cref{fig:result_2}, so there is some synergistic information in the different subspaces (\cref{fig:mi_pid_cifar} (c)).
Therefore, the ideal learner would build features from all subspaces, discarding redundant information and selecting the synergistic and unique parts.
It is natural to ask whether a deep network behaves like this.
We trained networks using the original images but tested them on images projected into different subspaces, e.g., low pass means that test images were projected into the low pass subspace of that index, and similarly for band pass and high pass.
Bands computed from PCA were used for CIFAR-10 classification, and the Fourier basis was used for ImageNet and other tasks.

Classification accuracy drops to chance for both CIFAR-10 and ImageNet if the network does not have access to lower bands; see~\cref{fig:result_4} (top).
For regression tasks, we plotted things differently and compared the error of a network trained on all bands but tested on a particular band to the error of a network trained on a particular band from~\cref{fig:result_2} (d).
If the error differential is large for a band, then the network trained on all the bands does not use information in that particular band.
\cref{fig:result_4} (bottom) shows that the error differential increases for band pass and high pass curves, i.e., networks predominantly use information in the head of the spectrum.
Unsurprisingly, the condition number of the optimization problem underlying these tasks is extremely large (\cref{fig:result_1}).
A large number of weight updates are necessary to fit the small amount of signal in higher bands.


\section{Discussion}
\label{s:discussion}

\paragraph{Implications for neuroscience.}

Natural data are statistically redundant~\citep{simoncelli2001natural}. The amplitude spectrum for luminance~\citep{atick1992does,field1987relations,van1996modelling} and color~\citep{burton1987color} falls off as $\sim 1/\abs{f}$ with frequency $f$ in natural images; this pattern is consistent across scales.
Spatio-temporal statistics follow similar trends~\citep{van1992theory,dong1995statistics,olshausen2000sparse}.
Edges~\citep{bell1997independent,guo2007primal}, colors~\citep{ruderman1998statistics,maloney1986evaluation}, contours~\citep{geisler2001edge} and textures~\citep{voorhees1988computing,tkavcik2010local}, all bear evidence of regularity in visual scenes.
Sounds, like images, also have self-similar power spectra~\citep{attias1997coding}, and obey sparsity-relevant patterns similar to those in vision~\citep{schwartz2001natural}.
Variations from these trends, and contextual differences that depend upon the task, are less studied~\citep{torralba2003statistics,hyvarinen2009natural}.
But more importantly, the role of this redundancy in the stimuli for a \emph{telos} (goal or purpose) is unclear~\cite{baker2021philosophical}.
We showed that redundancy in natural data goes beyond just the inputs. Typical tasks are also redundant functions of the input. The former is due to regularities in natural environments. While the latter is, perhaps, a property of tasks that biological organisms and machines \emph{chose to do}, i.e., tasks that are ecologically and economically useful.

Removing redundancy in the stimuli is fundamental to how the brain works~\citep{barlow1961possible,attneave1954some}.
There are circuits that implement whitening, low-pass filtering and compression~\citep{laughlin1981simple,atick1992does,barlow1989unsupervised}, sparse coding~\citep{barlow1972single,olshausen1997sparse,field1993scale,chechik2001group}, slow feature analysis~\citep{berkesSlowFeatureAnalysis2005,wiskottSlowFeatureAnalysis2002} etc.
But the efficient coding hypothesis could not be the complete story because there are three things at play: the environment, metabolic and architectural constraints imposed by the neural circuitry, and the task~\citep{simoncelli2001natural,fieldWhatGoalSensory1994,balasubramanian2009receptive}.
Although we know some principles to explain how neural circuits adapt to the
environment~\citep{ratliff2010retina,tkavcik2010local,balasubramanianHeterogeneityEfficiencyBrain2015}, in spite of intense activity, there is no clear understanding as to how tasks inform learned representations~\citep{rameshPictureSpaceTypical2023,mao2024training,rameshModelZooGrowing2022}.
Normative principles of representation learning~\citep{tishbyInformationBottleneckMethod1999,achille2018emergence,barlow2001redundancy,yerxa2023learning,yu2020learning} are, as yet, inadequate.

We showed that evidence for tasks is spread across different parts of the observation space.
If this evidence were nonredundant, tasks would be difficult to learn---both due to the condition number of the input spectrum and because of noise in the tail.
But because the evidence of the task is redundant, an organism, or a machine, need not be very careful as to which subspace it uses.
Learning, ontogenetic or over evolution, will refine neural circuitry and the representation, to improve performance.
There is also a second reason why this redundancy is useful.
Without it, a specific set of features would have to be learned for each task.
Changes in the task would severely impair the organism until the requisite features were learned.

\vspace*{-0.4em}
\paragraph{Implications for theoretical questions in deep learning.}
Consider logistic regression to fit labels $y_i = \sign\inner{w^*}{x_i}$ using a linear model $\hat y_i = \sign\inner{w}{x_i}$.
Even if the problem is strongly convex, gradient descent requires $\OO(\k \log(1/\e))$ iterations to reach $\e$-optimality where $\k$ is the condition number of $\sum_i x_i x_i^\top/n$~\citep{bottou2016optimization}. For all tasks in this paper, $\k > 10^6$.
Therefore, gradient descent evolves predominantly in the principal subspace of the input covariance matrix.
Spectral bias has been studied for linear models~\citep{hacohen2022principal}, kernel machines~\citep{yao2007early}, neural networks~\citep{rahaman2019spectral,tancik2020fourier} or the neural tangent kernel~\citep{cao2019towards}.
In short, effectively low-dimensional inputs result in ill-conditioned optimization problems~\citep{ma2017diving}.
Whitening risks amplifying noise, or spurious correlations from finite samples.

These works typically assume that the task needs all the features, or that it needs a few specific features.
But this does not explain why optimization problems underlying perception tasks can be solved so effectively in practice despite the large condition number.
If the task is redundant, say, $y_i = \sign\inner{w^*}{P x_i}$ for many matrices $P$ which project data into different subspaces, then since $\inner{w^*}{Px_i} =  \inner{P w^*}{x_i}$, there are many equivalent solutions and weight initializations with a large overlap with one of them.
Our results indicate that networks pick a particular projection $P$ that emphasizes the head of the input spectrum.

Alignment between labels and the principal subspace improves generalization for kernel methods~\citep{amini2022target} and two-layer networks~\citep{arora2019fine}.
If input data are effectively low-dimensional, then one can obtain analytical generalization bounds for deep networks~\citep{yang2021does,bartlett2020benign}, and also generalize well in practice~\citep{pope2021intrinsic,martin2021implicit}.
This also leads to some dramatic phenomena: networks of different architectures, training and regularization methods, evolve on extremely low-dimensional manifolds~\citep{mao2024training}.
Broadly, learning theoretic investigations do not consider the redundancy in the task.
Our observations suggest that this may be an important direction to investigate.

\vspace*{-0.4em}
\paragraph{Implications for practice.}
Reconstruction-based methods, e.g., masked auto-encoders (MAE), often take significantly longer to train before they generalize to downstream tasks.
\citet{balestriero2024learning} explained this by arguing that the tail of the spectrum is important for classification compared to the head of the spectrum.
Our numerical results are not at odds with theirs but our interpretation is different.
The different interpretations arise since the head and the tail of spectrum are defined using explained variance in \citet{balestriero2024learning}. 
For the explained variance of the tail to be as high as that of the head, one needs to use a very large number of eigenvectors---essentially all of them.
Our results paint a very different story.
The principal subspace is most important for the task.
But even a random subspace can predict the task remarkably well.
Therefore, while training MAEs is slow, the slow training and performance gap to contrastive learning may be for other reasons.


\section{Methods}
\label{s:methods}

\textbf{Principal components analysis (PCA), Fourier transform and the wavelet transform.}
The input to all of our image filters is a discretely-sampled image $x \in \reals^{d_1 \times d_2 \times d_3}$ with height $d_1$, width $d_2$ and $d_3$ channels.
For PCA, we flatten the image and subtract the mean over all images to form $X \in \reals^{d \times n}$ where $n$ is the number of examples and $d = d_1 d_2 d_3$.
We then form the sample covariance matrix given by $\S = X X^\top/n$.
PCA computes the eigen-decomposition $\S = U \L U^\top$ where $U$ is a matrix of eigenvectors and $\L$ is a diagonal matrix with eigenvalues $\l_1,\dots,\l_n$ along it.
If the eigenvalues are sorted, the explained variance of the subspace formed by the first $i$ eigenvectors is $1-(\sum_{j=1}^i \l_j)/(\sum_{j=1}^n \l_j)$.

We also consider the Fourier basis which represents images using an orthonormal basis of complex sinusoids of different frequencies
For natural images, pixel-pixel correlations are mostly only a function of the distance between them, i.e., $\S_{ij}$ only depends on $\abs{i-j}$.
Under this assumption, the eigenvectors of $\S$ are sinusoids~\citep{hyvarinen2009natural}; see~\cref{s:PCA}.
The  Fourier Transform is, hence, closely related to PCA.
We use the two-dimensional Discrete Fourier transform~\cite{briggs1995dft,oppenheim1999discrete} to project each channel of the image. The Discrete Fourier transform (DFT) of an image $x$ is
\[
    \hat x(\omega_1, \omega_2, c)=\frac{1}{d_1 d_2} \sum_{k=1}^{d_1}\sum_{l=1}^{d_2} x(k,l,c) \exp \cbr{-2\pi i \rbr{\frac{ \omega_1 k}{d_1} + \frac{\omega_2 l}{d_2}} },
\]
where $\omega_1, \omega_2$ are spatial frequencies. PCA and Fourier transform are a global statistic of the input image.
Wavelets can provide local statistics at different scales using a set of basis functions that have compact support in both frequency and space~\citep{mallat2008wavelet}. 
We use the discrete 2D wavelet transform for all our experiments with Daubechies 4 wavelets from the PyWavelet package.

\textbf{Slow feature analysis (SFA) for sounds.}
We used a cochlear model with 42 gammatone spectral filters followed by a temporal filter~\cite{Zhang2001-ac,lewicki2002efficient,tabibi2017investigating} to approximate the structure of audio inputs in our task to that of the auditory system.
Slow Feature Analysis (SFA~\citep{wiskottSlowFeatureAnalysis2002}) argues that higher-order information in a stimulus (e.g., identity) changes at a slower time scale than other fluctuations (e.g., acoustic features).
It has been shown to learn features similar to those in the V1 cortex~\cite{berkesSlowFeatureAnalysis2005}.
If the auditory stimulus, e.g., output of the cochlear model, is $x(t)$, features found by (linear) SFA are eigenvectors of the covariance matrix of the derivative $\dot x(t)$, arranged from the smallest eigenvalue to the largest, i.e., slowest to fastest.
SFA is equivalent to a PCA of the temporal derivative of a signal.
\cref{s:audition} provides more details.

\textbf{Shapley values~\citep{shapley1953value}}
calculate the improvement in the accuracy of a model with or without including a feature, averaged over all sets of other features~\citep{lundberg2017unified}.
This requires fitting exponentially many models.
But we can use an equivalent definition framed as a least squares problem with a linear constraint~\citep{covert2020improving} and optimize it using dataset sampling~\citep{ribeiro2016should} to estimate the average improvement in accuracy by including a particular subspace, formed by PCA/frequency/scales. The Shapley values sum up to the accuracy of the model when all features are included.

\textbf{Partial information decomposition (PID)~\citep{williams2010nonnegative}} was introduced to understand how the mutual information $I(X_1, X_2; Y)$ of two random variables $X_1, X_2$ with $Y$ can be decomposed into the redundant information $R$ (both $X_1$ and $X_2$ have), synergistic information $S$ (emerges only when both are available), and the unique information with respect to $Y$ for both variables, denoted by $U_1$ and $U_2$ respectively. By definition $I(X_1, X_2; Y) = R + S + U_1 + U_2$.%
\footnote{Consider two bits $X_1 \in \{0,1\}$ and $X_2 \in \{0,1\}$. If $\{0,1\} \ni Y = X_1 \text{ xor } X_2$, then $X_1$ and $X_2$ contain no redundant or unique information about $Y$. But there is 1 bit of synergistic information. If $X_1 = X_2 = Y$, then $X_1$ and $X_2$ contain 1 bit of redundant information for $Y$.}
We have
\begin{align}
    I(X_1; Y) = R + U_1, \text{ and } I(X_2; Y) = R + U_2. \label{eq:pid}
\end{align}
PID is not uniquely defined using these constraints. But it turns out that defining redundant information is enough.
\citet{williams2010nonnegative} provide a formula for $R$ but it is computationally intractable.
If $(X_1, X_2, Y)$ are jointly Gaussian, \citet{barrett2015exploration} showed that the definition of redundancy in \citet{williams2010nonnegative} reduces to 
\[
    R = \min \cbr{I(X_1; Y), I(X_2; Y)},
\]
which can be computed using the formula for the mutual information of Gaussian random variables. Using redundancy $R$, we can calculate $U_1, U_2$ and $S$ as well using~\cref{eq:pid}.

\textbf{Estimating mutual information.}
Calculating the mutual information $I(X;Y) = H(Y) - H(Y \mid X)$ from samples requires a numerical estimate of the entropies.
The Kraskov estimator~\citep{kraskov2004estimating} uses a $k$-nearest neighbor based estimate of the entropy and we additionally exploit the identity $I(X; Y) = H(X) + \sum_y P(Y=y) H(X \mid Y=y)$ for discrete $Y$.
Upon this, one can implement a local non-uniformity correction (LNC) term~\citep{gao2015efficient}, which is important when variables are strongly correlated.
\citet{belghazi2018mine} developed an estimator called MINE that uses a neural network to optimize the Donsker-Varadhan inequality.
In general, estimating mutual information in high dimensions is difficult~\citep{czyz2024beyond}.
We only estimate it for labels and 10-dimensional PCA subspaces, so we expect our estimates to be reliable.

\section{Acknowledgement}

PC would like to acknowledge funds provided by the National Science Foundation (IIS-2145164, CCF-2212519), the Office of Naval Research (N00014-22-1-2255), the National Science Foundation and DoD OUSD (R\&E) under Cooperative Agreement PHY-2229929 (The NSF AI Institute for Artificial and Natural Intelligence) and cloud computing credits from Amazon Web Services.
KD would like to acknowledge the National Science Foundation (NSF FRR 2220868, NSF IIS-RI 2212433) and the Office of Naval Research (ONR N00014-22-1-2677).  VB and RWD were supported in part by NSF under CISE 2212519, and by the NSF and DoD OUSD (R\&E) undert PHY-2229929 (The NSF AI Institute for Artificial and Natural Intelligence). VB was also supported by the Eastman Professorship at Balliol College, University of Oxford


\setlength{\bibsep}{1.7ex}
\clearpage
\bibliographystyle{plainnat}
\bibliography{bib/ref,bib/pratik,bib/RWDRef}

\clearpage

\appendix
\renewcommand\thefigure{A.\arabic{figure}}
\renewcommand\thetable{A.\arabic{table}}
\setcounter{figure}{0}
\setcounter{table}{0}
\renewcommand{\figurename}{Figure}
\renewcommand{\tablename}{Table}

\section{Experimental Setup}
\label{s:setup}

In this section, we provide details regarding the datasets, network architectures, hyper-parameters and discuss how we apply different low pass, high pass and band pass filters. The experiments in this paper required about 1000 GPU-hours and span a diverse set of perception tasks and even an additional auditory task.

\subsection{Datasets}
We examine various complex perception tasks, including classification, semantic segmentation, optical flow, and depth/disparity prediction.  In this work, we use CIFAR-10 for classification, ImageNet for classification, the Cityscapes dataset for semantic segmentation and disparity prediction\cite{cityscapes}, the ADE20K for semantic segmentation\cite{ade20k} and an augmented version of the M3ED dataset\cite{m3ed}. We feature complex real scenes beyond just classification datasets. For example, the M3ED dataset has many natural scenes, including cars driving in urban, forest, and daytime and nighttime conditions.

\subsection{Projecting data onto different subspaces}\label{s:app:index}

A key operation in this work is projecting input data onto different subspaces using PCA, Fourier and wavelet bases.

\paragraph{Indices for the bases.} For each basis, we define an index that determines the ordering of the basis elements. The indices are defined as follows:
\begin{enumerate}[(1),nosep]
    \item For PCA, we order the basis elements by the descending order of the eigenvalues. The basis element with index 5 is the eigenvector corresponding to the 5th largest eigenvalue.
    \item For the Fourier basis, the basis elements are ordered by the increasing radial frequency. The basis elements with index $i$, contains all spatial frequencies $(\omega_1, \omega_2)$ such that
        \[ i \leq  \sqrt{\omega_1^2 + \omega_2^2} < i+1. \]
    \item For wavelets, the basis elements are ordered by increasing scales. The small scales capture low-frequency information, while the larger scales capture high frequency information. For example, a 2-level wavelet decomposition of an image contains 7 coefficients. The approximation coefficient (cA) represents the smallest scale. Each level has 3 detail coefficients (cH, cV, cD) representing the horizontal, vertical, and diagonal details, respectively. The ordering of the wavelet coefficients by scale(small-large) is (cA2, cH2, cV2, cD2, cH1, cV1, cD1). Hence the basis elements with index 5 correspond to the coefficients cH1.
\end{enumerate}

\paragraph{Bands.} We use the ordering of the basis elements to define different "bands" of the basis. Low pass and high pass filters consider basis elements that are lesser than or greater than a certain index. Band pass filters consider basis elements between two indices. We train networks after projecting the input data onto different subspaces defined by the low pass, high pass, and band pass filters. After projecting onto these subspaces, we transform the input back to pixel space and train the network.

We perform experiments with PCA, Fourier and Wavelet bases for CIFAR-10. However, computing PCA on datasets as large as ImageNet is computationally difficult, so we study the larger datasets using the Fourier or wavelet bases. In~\cref{fig:result_1}, we find a strong correspondence between the two, which suggests that one should get similar results using the PCA, Fourier and wavelet bases.

\subsection{Neural network architectures and training procedure}\label{ss:network}

\begin{table}[ht!]
\centering
\renewcommand{\arraystretch}{1.25}
\resizebox{\linewidth}{!}{
\footnotesize
\begin{tabular}{l rrrr rrr rr}
\toprule
\textbf{Dataset} & \textbf{$e_{tot}$} & \textbf{$\lambda$} & \textbf{$\lambda_{max}$} & \textbf{Batch Size} & \multicolumn{3}{c}{\textbf{Channels}}& \multicolumn{2}{c}{\textbf{\# Samples}}\\
&&  &  & & \textbf{In} & \textbf{Out} & \textbf{Base} & \textbf{Train} & \textbf{Test} \\
\midrule
CS Depth & 50 & 0.0001 & 0.005 & 128 & 6 & 1 & 64 & 22972 & 500 \\
M3ED Depth & 40 & 0.001 & 0.01 & 128 & 6 & 1 & 64 & 34255 & 15020 \\
M3ED Flow & 40 & 0.001 & 0.01 & 128 & 6 & 2 & 32 & 34255 & 15020 \\
M3ED Sem & 60 & 0.005 & 0.0001 & 128 & 3 & 20 & 64 & 54793 & 13822 \\
Cityscapes Sem & 40 & 0.0005 & 0.01 & 128 & 3 & 20 & 64 & 22973 & 500 \\
ADE20k Sem & 60 & 0.0001 & 0.01 & 128 & 3 & 150 & 64 & 20210 & 2000 \\
\bottomrule\\
\end{tabular}
}
\caption{Hyperparameters and dataset sizes for different datasets}
\label{tab:hyperparameters}
\end{table}

In our experiments, we apply various low-pass/high-pass/band-pass filters and train networks based on the results of these filters. We then measure the test error (1-Accuracy, for classification and segmentation), the mean average error (for depth estimation), or the average endpoint error (optical flow). The different filters can tell us which subspaces of the input have information relevant for classification--for example, if a high-pass filter results in chance accuracy on a classification task, then this suggests that there is little information for the task present in the tail of the spectrum. However, our results suggest otherwise, the entire spectrum performs well on our tasks.

\paragraph{Classification.} For CIFAR-10, we consider the Wide-Resnet architecture\citep{zagoruyko2016wide} with 16 layers and 4 blocks of sizes 16, 64, 128 and 256. We normalize the images by the mean and standard deviation but do not apply any augmentations during training. The models are trained for 100 epochs using stochastic gradient descent with Nesterov momentum (of 0.9) and use a learning rate of 0.05 with a batch size of 64 and weight decay of $5 \times 10^{-5}$. The networks are trained to optimize the cross-entropy loss.

The ImageNet models are trained using the Resnet-50~\citep{he2016deep} architecture with the pooling layers replaced with BlurPool~\citep{zhang2019making}. To speedup training, we use FFCV~\citep{Leclerc_2023_CVPR}, which is a data-loading library optimized to load and perform augmentations on the training data quickly. We train the models for 40 epochs with progressive resizing – the image size is increased from 160
to 224 between the epochs 29 and 34. The projection operation (band pass, low pass, or high pass) is applied to an image of size 256 regardless of the stage of training in progressive resizing. Models are trained on 4 GPUs with a batch-size of 512. We use random-resized crop and random horizontal flips as the two augmentations and label smoothing parameter set to 0.1. We use stochastic gradient descent as the optimizer with the learning rate schedule that is annealed linearly. We use a weight decay of 0.0001 but do not apply it to the batch norm parameters.

\paragraph{Semantic segmentation, Optical Flow and Depth prediction.} All images in this study were down-sampled to 200$\times$200, except for M3ED semantic segmentation which is $180\times 200$. We apply filters starting from 10- half the minimum resolution for the various low-pass and high-pass filters. As for band-pass filters, we again apply them using the same interval with a width of 10 for all dense perception experiments. For each experiment, one of these filters is applied to the input image spectra and used as input for the model of our task. We tested most of our models on a U-Net~\cite{unet} architecture with 3 layers with a 2x downsampling and channel multiplier.  These backbones contain downsampling/upsampling operations; therefore, to ensure a valid sized output, we pad our input evenly and crop the output for loss calculation. We train all models using supervised training with backpropagation using  AdamW~\cite{adamw} with OneCycleLR~\cite{onecycle} policy with an initial learning rate of $\lambda$ and a maximum learning rate of $\lambda_{max}$ with a batch size of 128 for $e_{tot}$ epochs and fp16/bf16 weights.

Now, we will review the respective input and training loss used for each task. The task of optical flow takes in two sequential in time RGB images($6\times M\times N$) and regresses optical flow($2\times M \times N$) with 2 channels corresponding to the flow in the x and y directions. The loss function for optical flow is the robust \cite{charbonnier} comparing the ground truth and estimated optical flow along with a second-order smoothness loss weighted with a penalty of $\beta=0.5$.  In both cases, our depth prediction takes in two frames, for M3ED this corresponds to two sequential frames and for Cityscapes this corresponds to a left and right camera(in this case we predict disparity).  We again use the robust \cite{charbonnier} comparing the ground truth and estimated depth. Finally, our semantic segmentation task takes as input one RGB image ($3\times M\times N$) and regresses a distribution of semantic classes($K\times M \times N$), for classes K. The loss function for semantic segmentation was the cross entropy loss between the predicted class distribution and actual class distribution per pixel. These tasks are masked based on whether ground truth labels are available for this output pixel. 

\subsection{Random bands used in experiments}
\label{app:rnd_bands}

\begin{table}[H]
    \centering
    \begin{tabular}{lcc}
        \toprule
        & Center Frequency & Width \\
        \midrule
        Depth & 5, 30, 50, 80 & 2 \\
        & 10, 50, 80 & 5 \\
        & 5, 45, 75 & 5 \\
        & 5, 40, 60, 80 & 2 \\
        & 15, 45, 85 & 5 \\
        \midrule
        Semantic Segmentation & 5, 30, 50, 80 & 2 \\
        & 10, 50, 80 & 5 \\
        & 5, 45, 75 & 5 \\
        & 5, 40, 60, 80 & 2 \\
        & 15, 45, 85 & 5 \\
        \midrule
        Optical Flow & 25, 40, 60, 80 & [16, 8, 4, 2] \\
        & 5, 40, 60, 80 & 2 \\
        & 30, 55, 70, 90 & 2 \\
        & 5, 10, 30, 50, 70, 90 & 2 \\
        & 30, 60, 80, 93 & [16, 8, 4, 2] \\
        & 17, 50, 70, 90 & [16, 8, 4, 2] \\
        & 10, 30, 50, 70, 90 & 6 \\
        & 5, 30, 50, 80 & 2 \\
        & 17, 50, 70, 90 & 5 \\
        \bottomrule
    \end{tabular}
    \caption{Random bands used in~\cref{fig:random_pass_band}. For each of the experiments, we list the center frequency used and the corresponding width of that pass band. For experiments with multiple widths, each width is used for the corresponding sequentially ordered center frequency.  }
    \label{tab:rnd_bnds_info}
\end{table}

\subsection{Experimental setup for audition tasks}
\label{s:audition}
We used a cochlear model with two stages of filters: 42 gammatone spectral filters followed by a temporal filter~\cite{Zhang2001-ac,lewicki2002efficient,tabibi2017investigating} to approximate the structure of audio inputs in our task to that of the auditory system.
These gammatone filters had center frequencies between 22.9 Hz to 20208 Hz, which covered the frequency range of all vocalizations studied. The temporal filter was implemented as a difference of two kernels of the form:
\[
    g(n)=a n^m e^{-bn},
\]
in which $n$ is in units of samples and $a$, $b$, and $m$ are filter parameters.
The temporal filter was created by taking the difference between $g_1$ and $g_2$ with the following parameters: $g_1$: $a = 1.5$, $b = 0.04$, and $m = 2$; and $g_2$: $a = 1$, $b = 0.036$, and $m = 2$.
This parameter set accounts for some key aspects of cochlear temporal processing \cite{lyon2010history,tabibi2017investigating}.
Each filter output was normalized to have zero mean and unit standard deviation.

The conceptual underpinning of Slow Feature Analysis (SFA) is the “slowness principle”.
This principle hypothesizes that higher-order information in a stimulus (e.g., stimulus identity) changes at a slower time scale than other lower-order fluctuations (e.g., acoustic features).
SFA learns the slow features in a stimulus through unsupervised learning.
The formulation for linear SFA is described in the following equations:
\[
    \aed{
        \reals^m \ni y(t) &= \hat w^\top x(t)\\
        \hat w &= \argmin_{w \in \reals^{d \times m}} \agr{\norm{y(t) - y(t-1)}^2_2};
    }
\]
where $x(t) \in \reals^d$ is the auditory stimulus (output of the cochlear model), $w \in \reals^{d \times m}$ is the set of slow features and $y(t)$ is the projection of the stimulus into the feature space. The notation $\agr{}$ denotes an expectation over time.
Slowness is realized by finding $m$ features that minimize the average squared temporal difference of $y(t)$.
This effectively assigns weights to the original stimulus features to generate an output signal $y(t)$ that changes most slowly across time. To obtain non-trivial solutions, one imposes the constraint that the SFA features $y(t)$ have zero-mean and unit variance across time, and are uncorrelated with each other.
Features found by SFA are the eigenvectors of the covariance matrix of the derivative of the auditory stimulus $x(t)$, arranged from the smallest eigenvalue to the largest, i.e., slowest to fastest.

We use a multi-layer perceptron (MLP) to discriminate between vocalization pairs. The input layer has dimensionality equal to the number of SFA features, the hidden layer with ReLU nonlinearities has 5 neurons and a single output neuron to discriminate between two categories.
Adam is used to fit this MLP with a 50-50 stratified split of the train and test data; we used bootstrap (5 runs) to report test errors.

\section{Additional Results}

\subsection{Auto-encoder Training}
To understand what reconstruction losses actually capture in standard autoencoder representation learning, we conducted experiments to investigate this question. For these experiments, we trained a 3-layer encoder consisting of linear layers with ReLU activation and a similar decoder with no activation function in the output layer. This autoencoder was trained to reconstruct the input data using mean squared error (MSE) loss with backpropagation. We used the AdamW optimizer with a learning rate of 1e-3 for 200 epochs. Our network architecture had the following dimensions: it started with the original dimensionality of 3072 (corresponding to a CIFAR image), then reduced it to 3200, maintained the same dimensionality in the next layer (3200), and finally encoded the data into a latent space of size 1600. 

\begin{figure}[bt]
    \centering
    \includegraphics[width=\linewidth]{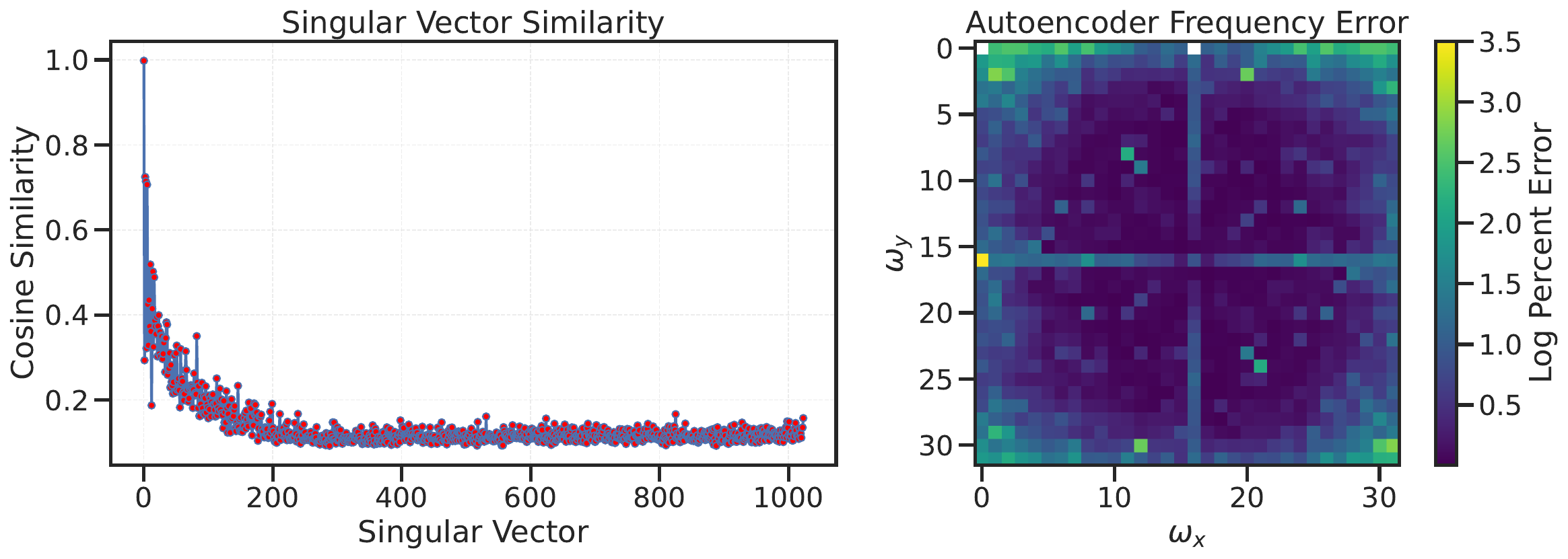}
    \caption{We trained a 3-layer auto-encoder to inspect the original and reconstructed data distribution properties. We look at: 1) the similarity of the original data distributions eigenvectors and the reconstructed. 2)  The error in the frequency domain of the residual of these two, showing the largest error regions are in the high frequency area. }
    \label{fig:autoencoder_learn}
\end{figure}

After training, we obtained the reconstructions produced by the autoencoder for all the original input images. We then analyzed the singular vectors of these reconstructions. Figure~\ref{fig:autoencoder_learn} illustrates the similarity between the singular vectors of the original data distribution and the closest singular vectors in the reconstructed dataset. Furthermore, we examined the difference between the mean of the reconstructed images and the original images in the frequency domain. We observed that the largest error regions and most dissimilar vectors were concentrated in the high-frequency area. Based on these results, we infer that the network primarily utilizes low-frequency information and does not effectively encode high-frequency information.

\subsection{Additional Perception Filter Experiments }
In addition to the band-pass, low-pass, and high-pass filtering experiments in the main work, we conducted further experiments to investigate the impact of these factors on task performance.

The first experiment involved applying a sinc mask to the frequency distribution of the original input data. Increasing the sinc frequency led to a consistent decrease in the average endpoint error for optical flow estimation.

The next experiment examined the effect of normalizing the input distribution to a range of -1-1 across all our experiments. We observed that the trends remained unchanged compared to the unnormalized input. This is likely due to the batch normalization layers employed in the U-Net architecture.

\begin{figure}[H]
    \centering
    \includegraphics[width=\linewidth]{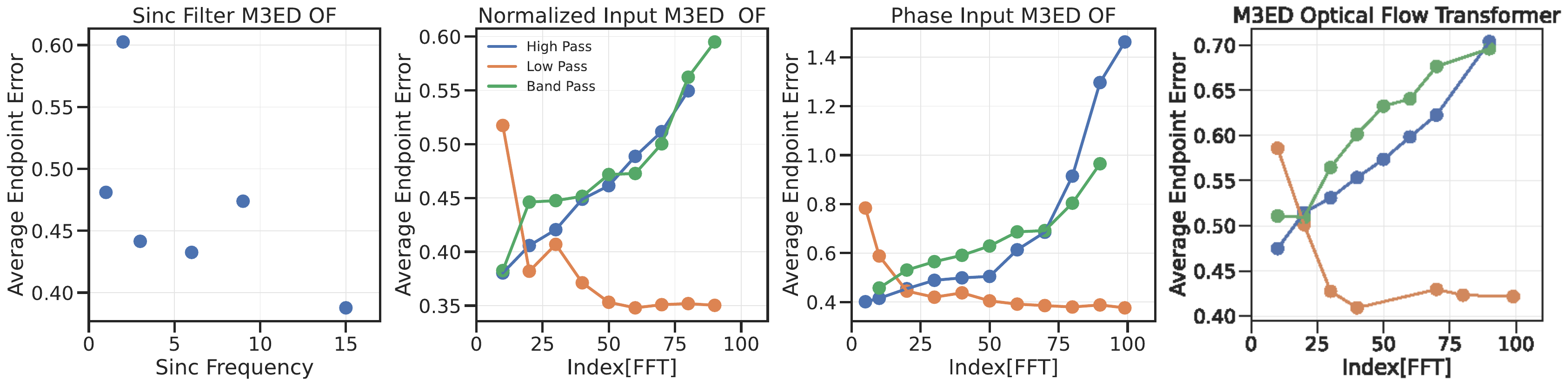}
    \caption{In addition to the standard set of baselines shown, we also created some additional results on the M3ED optical flow dataset. 1) Sinc filter that nulls out bands according to sinc frequency. 2) Normalize the input after filtering to range from -1-1., 3) Create a mask just on the phase distribution of the FFT(note this mask was a hard 0-1, mask as opposed to Butterworth filter) 4) Efficient Former V2\cite{efficienttransformer} was used for the standard high pass, low pass and band pass in this work. }
    \label{fig:perc_extra}
\end{figure}

In another experiment, we investigated masking out the phase information of the original input image instead of the amplitude in the frequency spectrum. This approach was inspired by research in psychophysics and signal processing~\cite{ImpOfPhase, Thomson2000HumanST, Wichmann2006PhaseNAHuman}, which highlights the impact of phase statistics on image perception. The results of this experiment mirrored those of the original M3ED experiments, showing that performance improves with a larger portion of the spectrum preserved and that low-frequency information leads to better performance compared to high-frequency information.

For our final set of experiments, we aimed to investigate changes in performance by replacing the U-Net architecture with the EfficientFormer V2 model~\cite{efficienttransformer}. The overall trends in the results remained consistent with those obtained using the U-Net model.

\subsection{Wavelet Results}
For the perception tasks, we primarily focused on the Fourier basis, which considers how a linear basis affects task performance by examining images at a global scale. However, another approach to redundancy reduction involves removing it across various spatial scales. Image statistics are generally believed to be invariant to scale changes. This suggests that we can investigate the impact of local redundancy removal on more global levels of redundancy \cite{,torralba2003statistics,Zoran2013NaturalIS}. Therefore, to explore how redundancy changes at different spatial scales, we employ a wavelet transform \cite{mallat2008wavelet} on the 2D image and analyze how these frequency bands influence our algorithm's performance.
\begin{figure}[H]
    \centering
    \includegraphics[width=0.45\linewidth]{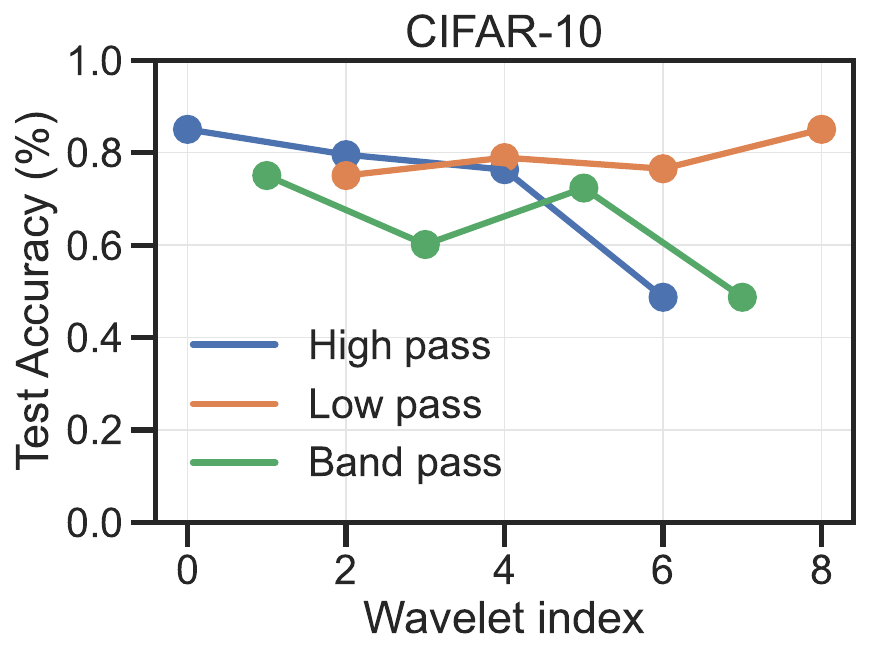}
    \caption{We performed the wavelet experiments described on the CIFAR10 dataset and again found that using all the information leads to the best performance, but we also perform remarkably well by using just a subset of the scales.}
    \label{fig:wavelet_cifar}
\end{figure}
\begin{figure}[H]
    \centering
    \includegraphics[width=0.9\linewidth]{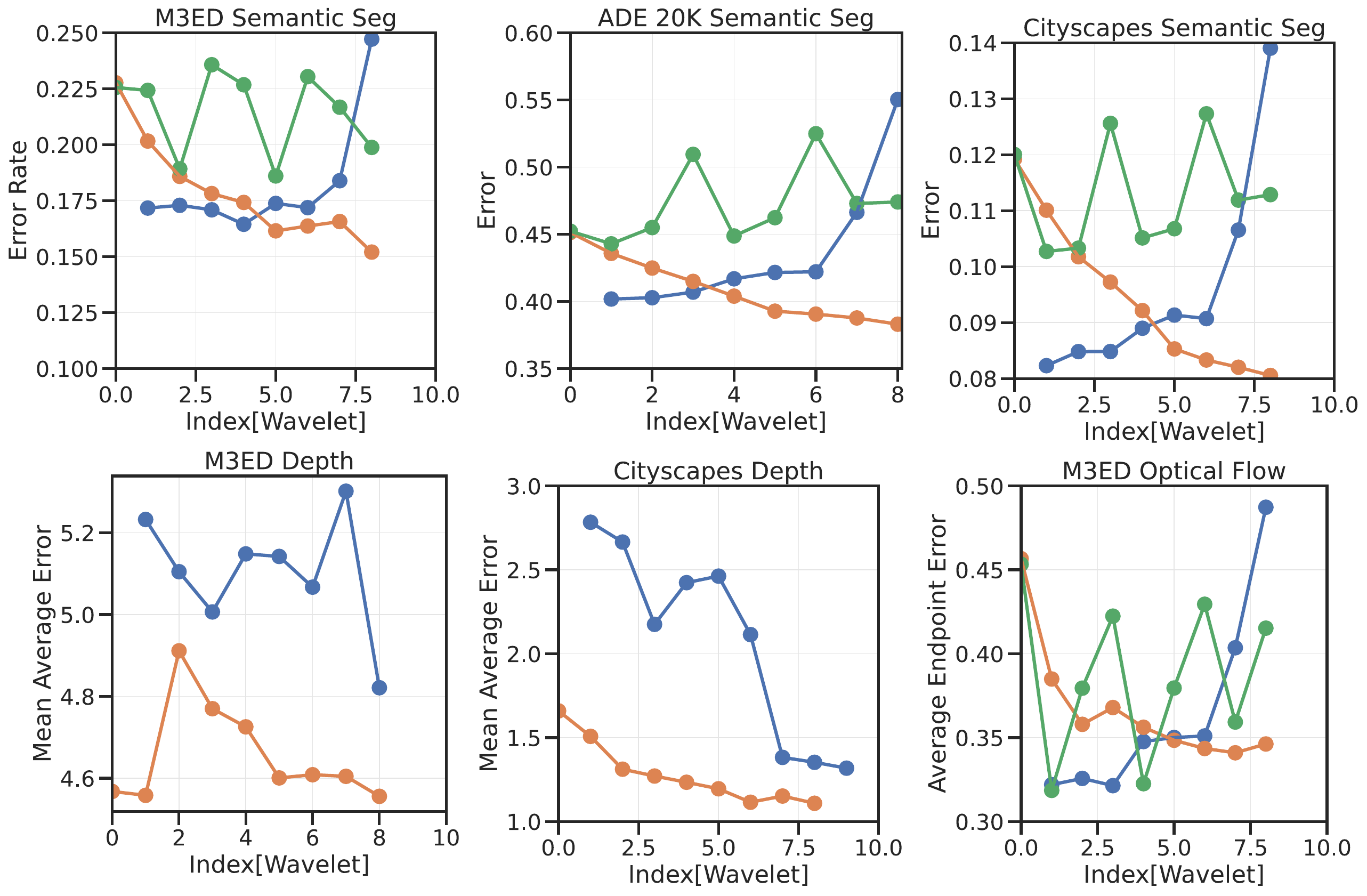}
    \caption{We performed the wavelet experiments described in the paper across various perception tasks. We find, in general, using all the information leads to the best performance, but across both high and low frequency, total additional information increases negligibly. }
    \label{fig:wavelet_perception}
\end{figure}

\subsection{Understanding redundancy in feature space}

\begin{figure}[H]
    \centering
    \includegraphics[width=0.4\linewidth]{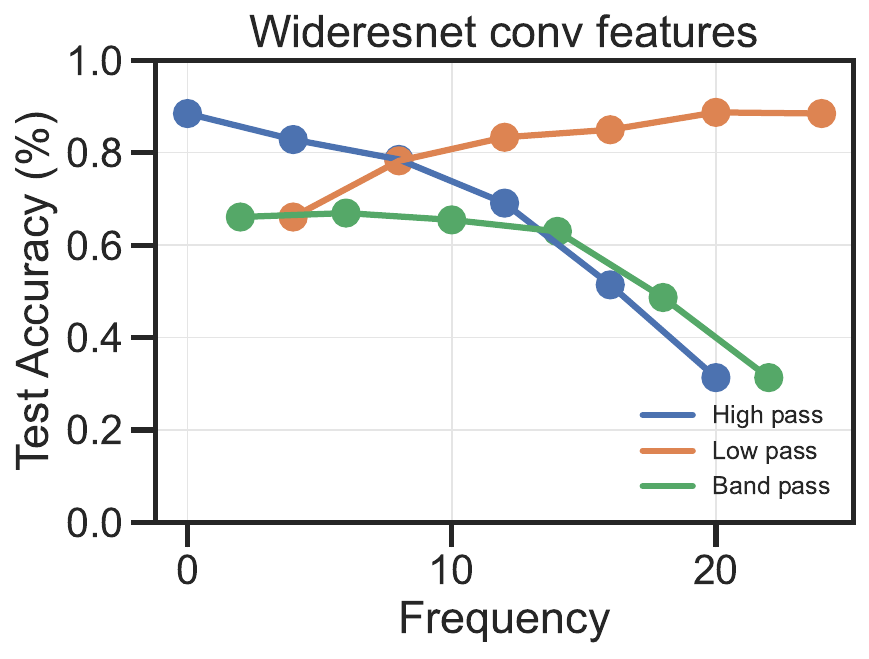}
    \includegraphics[width=0.4\linewidth]{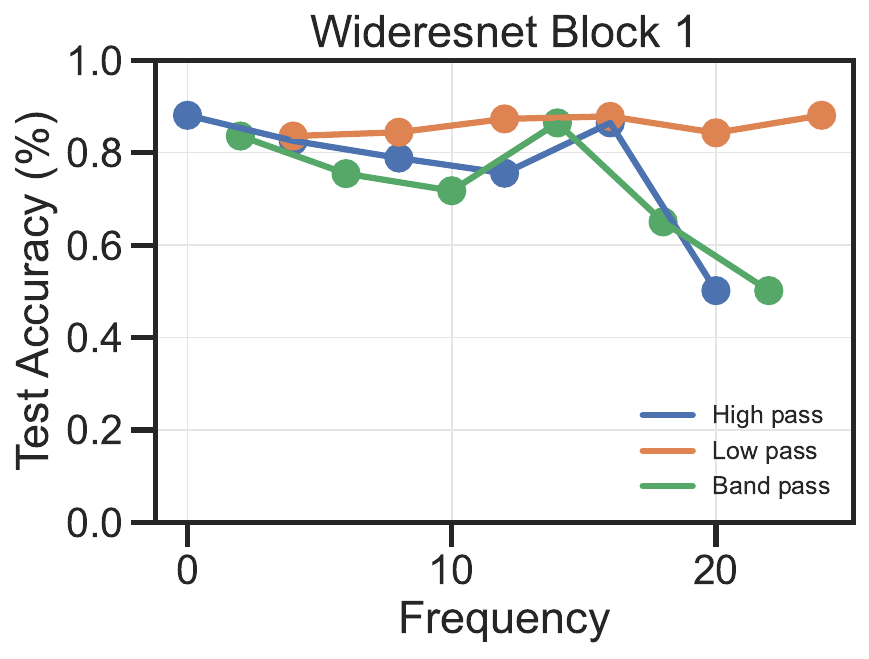}
    \includegraphics[width=0.4\linewidth]{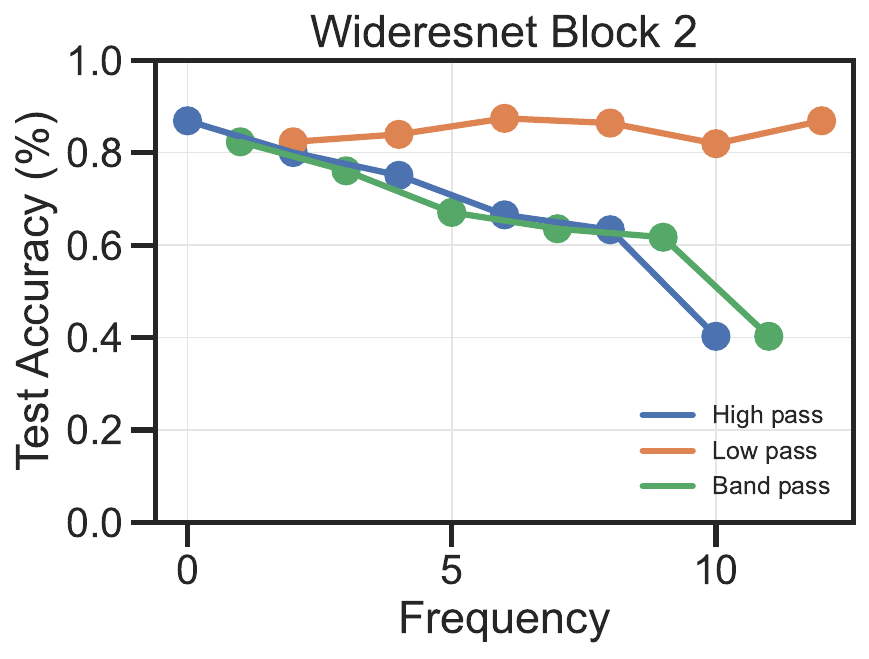}
    \includegraphics[width=0.4\linewidth]{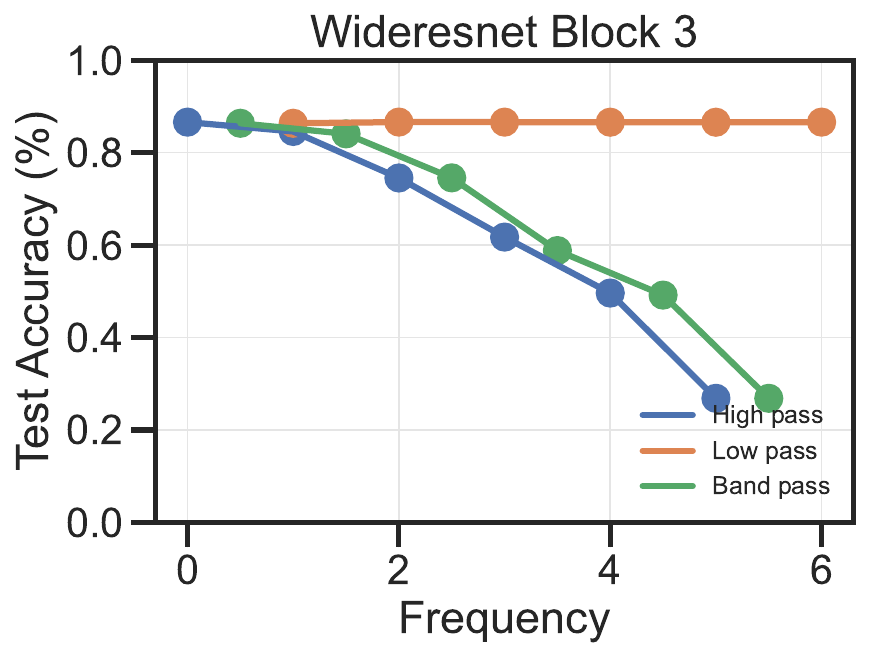}
    \caption{There is also redundancy in the learned features of a trained network.
    The experimental setup of this figure is identical to that of~\cref{fig:result_4} (top) except that instead of creating subspaces using PCA, Fourier and wavelet bases computed from raw pixels, we create these bands using the features from different intermediate layers of the trained network.
    In other words, there is redundancy in the original input data because many subspaces are predictive of the task. However, the network does not completely remove this redundancy when it learns the features.
    }
    \label{fig:app:feature_space}
\end{figure}

\subsection{Partial Information Decomposition}

\begin{figure}[H]
    \centering
    \includegraphics[width=0.3\linewidth]{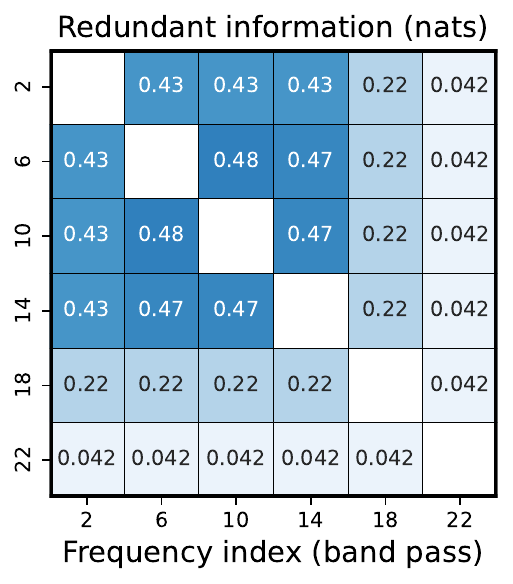}
    \includegraphics[width=0.3\linewidth]{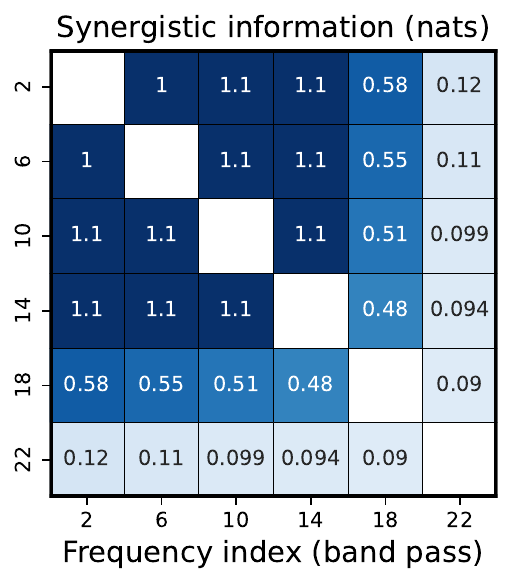}
    \includegraphics[width=0.3\linewidth]{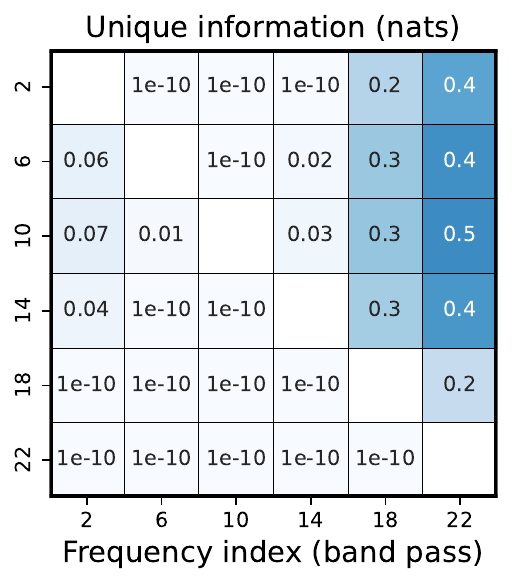}
    \includegraphics[width=0.3\linewidth]{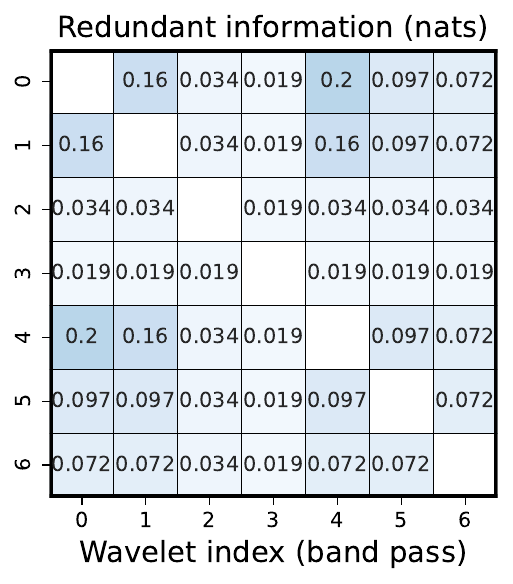}
    \includegraphics[width=0.3\linewidth]{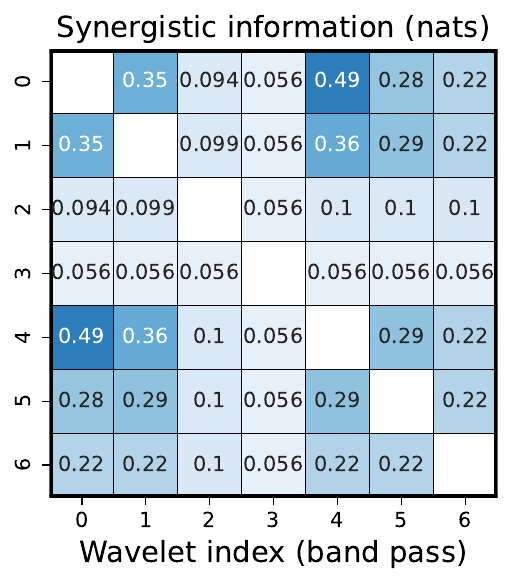}
    \includegraphics[width=0.3\linewidth]{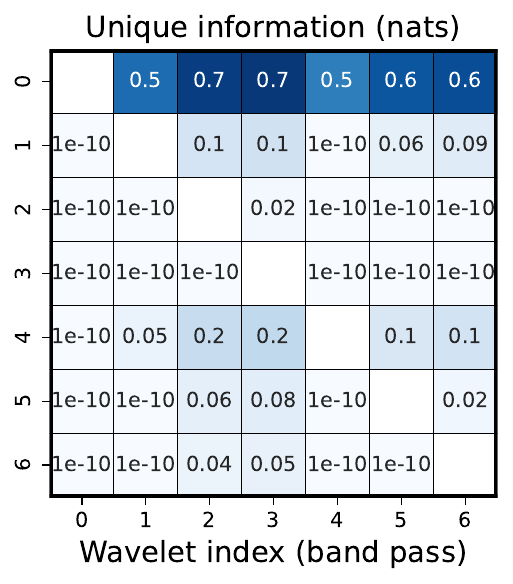}
    \caption{Different bands have high amounts of redundant and synergistic information even with frequencies and wavelets for CIFAR-10. This corroborates the result in~\cref{fig:mi_pid_cifar}. It also suggests that our technique for calculating mutual information is reliable; we see that redundant and synergistic information are large while unique information is small. Note that the numbers here are in log-scale, while the ones in the main paper are in nats.}
    \label{fig:pid_cifar_2}
\end{figure}

\subsection{Projections of images onto different subspaces}

We project an image of CIFAR-10 onto different bands of eigenvectors, frequencies and wavelets. While the eigenvectors, frequencies, and scales with larger indexes are imperceptible to the human eye, deep networks achieve non-trivial accuracy when trained on these bands.

\begin{figure}[H]
    \centering
    \includegraphics[width=\linewidth]{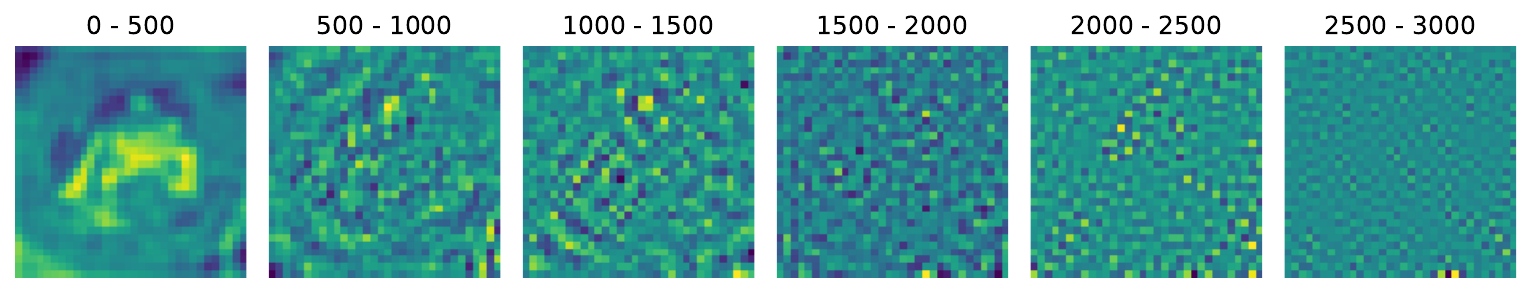}
    \caption{Image of Frog from CIFAR-10 when projected into different PCA subspaces.}
    \label{fig:projections_svd}
\end{figure}
\begin{figure}[H]
    \centering
    \includegraphics[width=\linewidth]{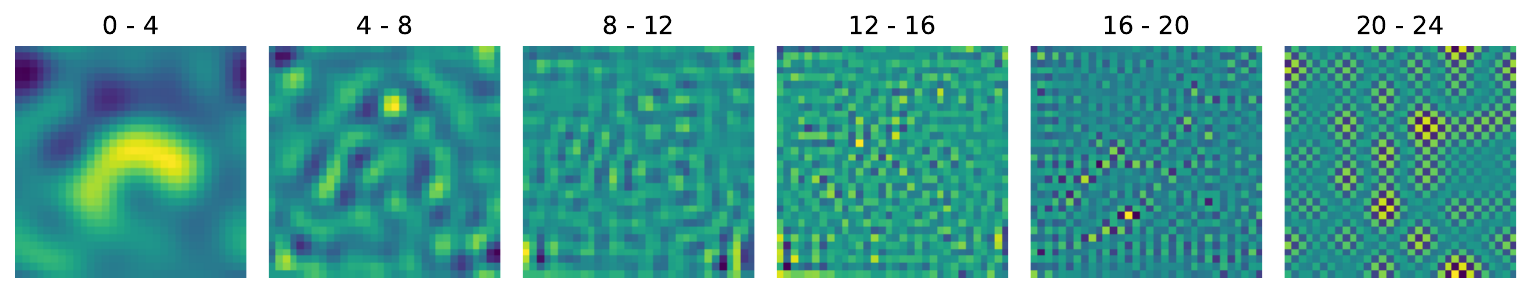}
    \caption{Image of Frog from CIFAR-10 image when projected into different radial frequency bands.}
    \label{fig:projections_fft}
\end{figure}
\begin{figure}[H]
    \centering
    \includegraphics[width=\linewidth]{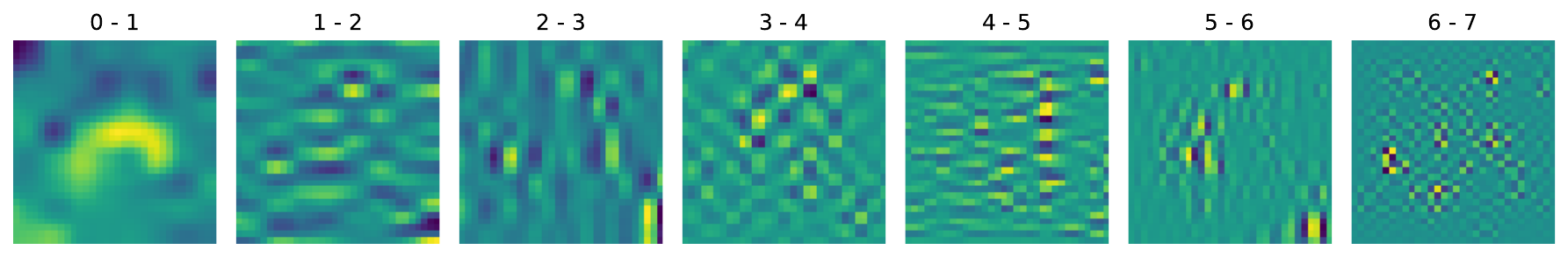}
    \caption{Image of Frog from CIFAR-10 when projected into different wavelet scale bands.}
    \label{fig:projections_wav}
\end{figure}

\subsection{Understanding bands of eigenvectors, Fourier and wavelet bases}

We plot the number of features and the power of each band for CIFAR-10.

\begin{figure}[H]
    \centering
    \includegraphics[width=0.325\linewidth]{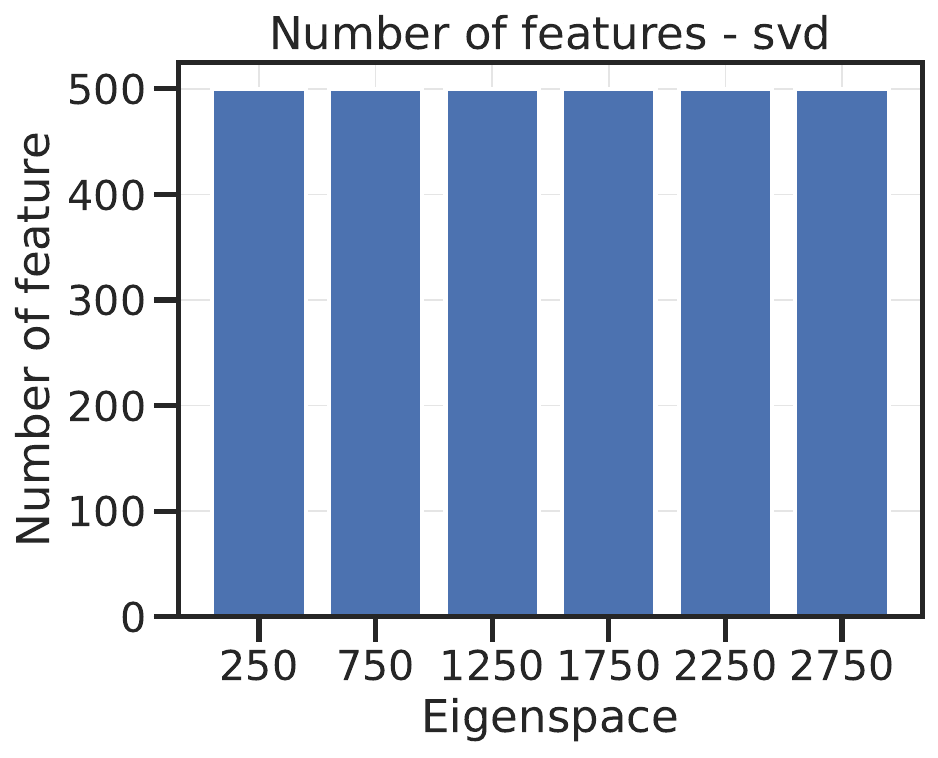}
    \includegraphics[width=0.325\linewidth]{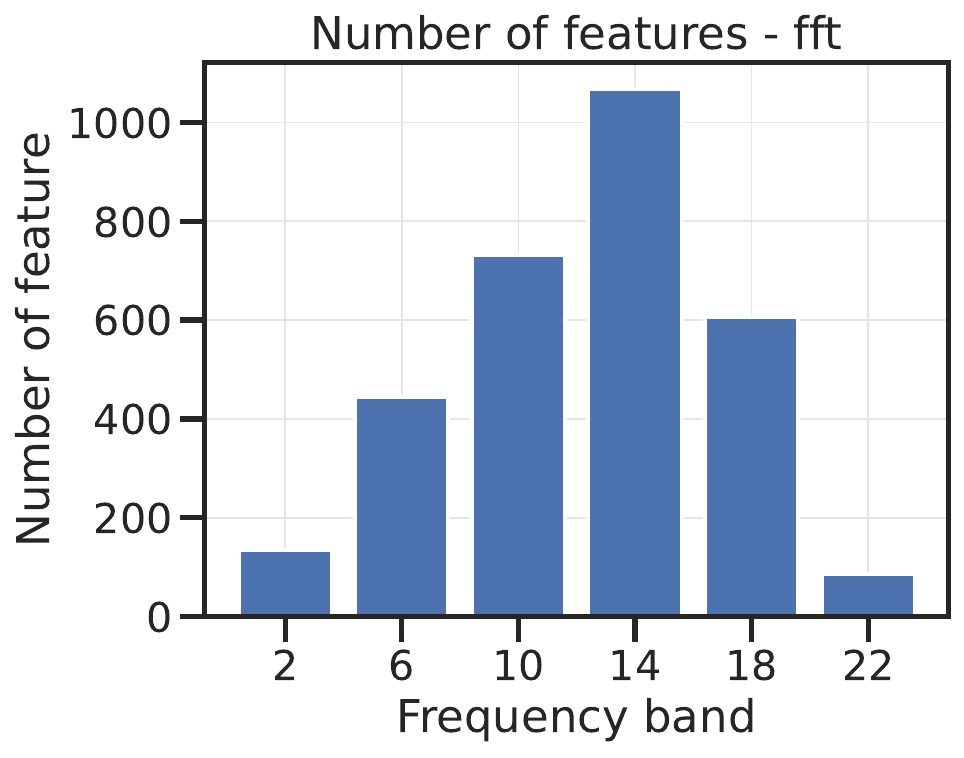}
    \includegraphics[width=0.325\linewidth]{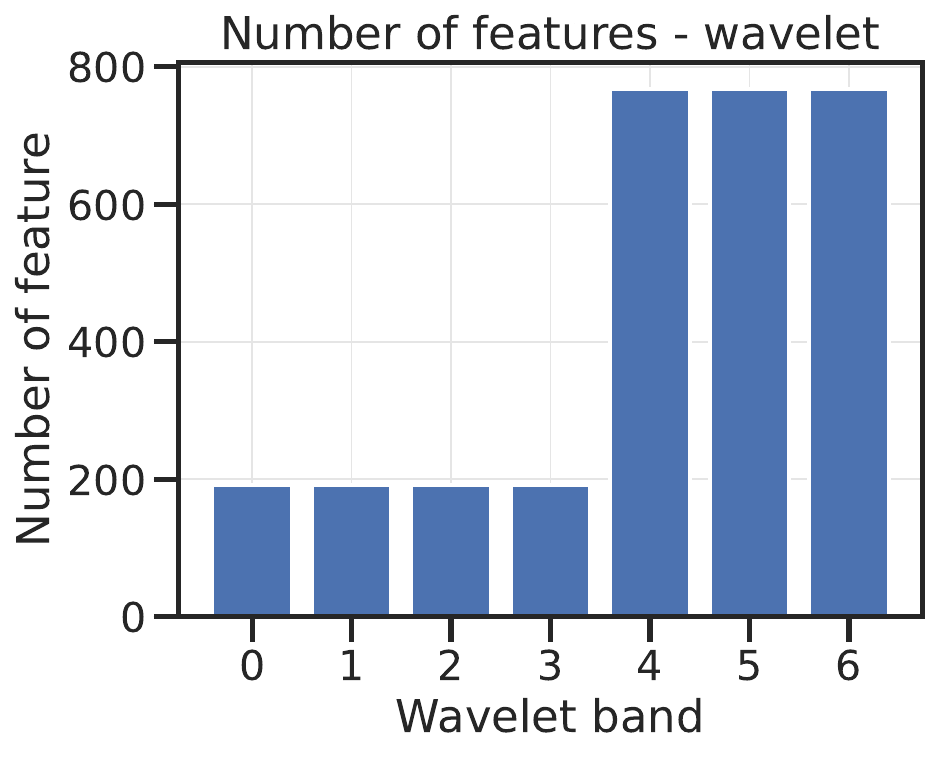}
    \caption{Number of features in different bands of the input.}
\end{figure}

\begin{figure}[H]
    \centering
    \includegraphics[width=0.325\linewidth]{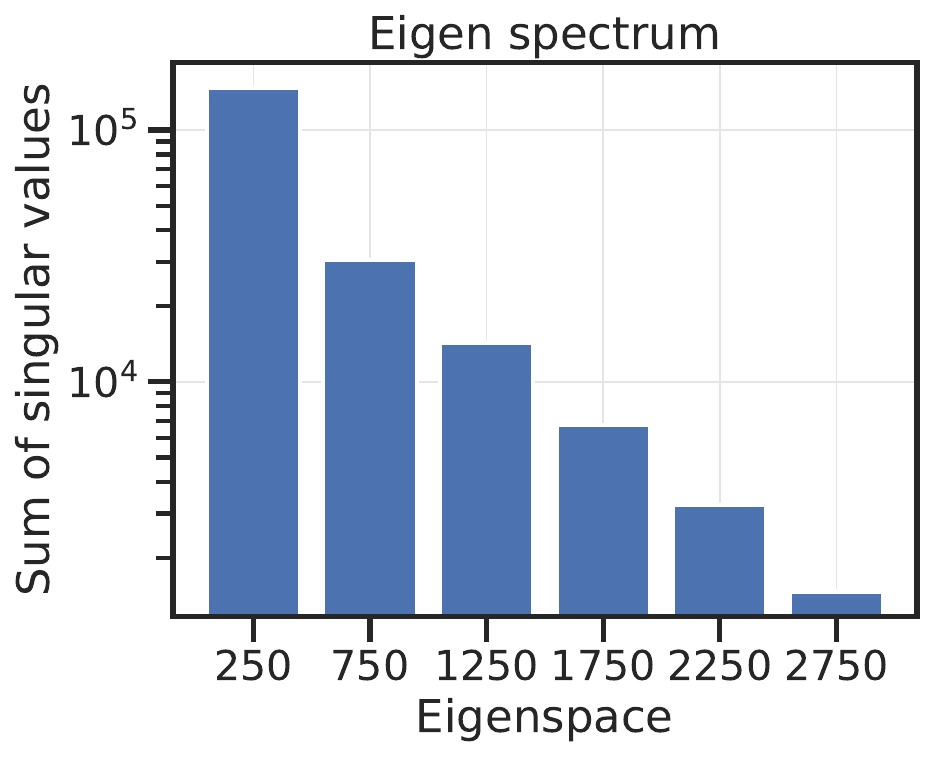}
    \includegraphics[width=0.325\linewidth]{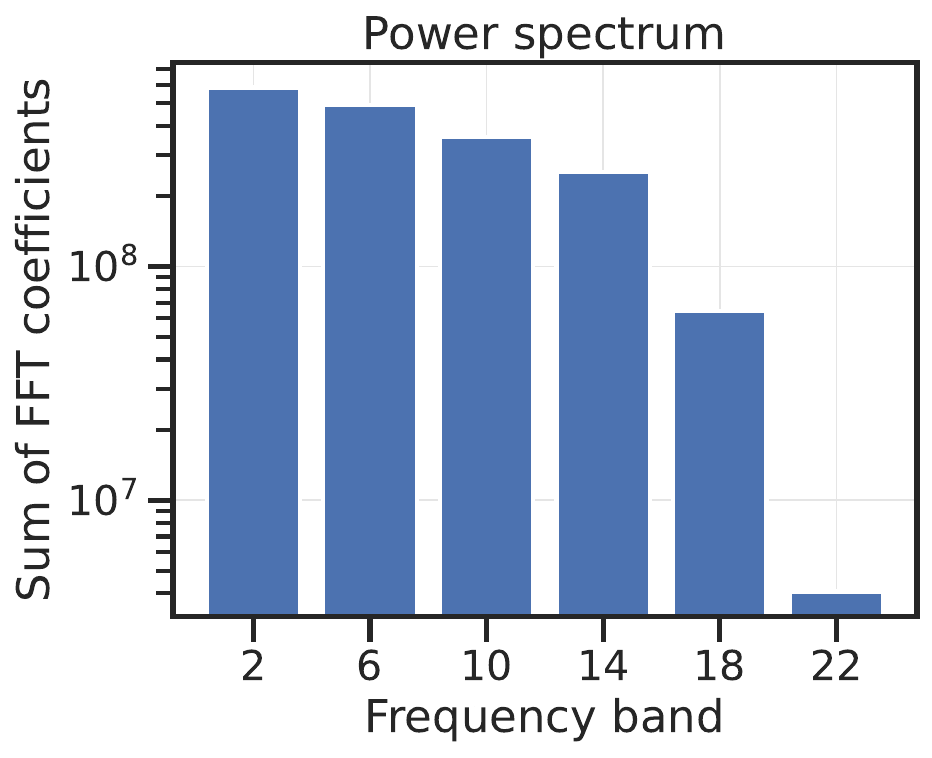}
    \includegraphics[width=0.325\linewidth]{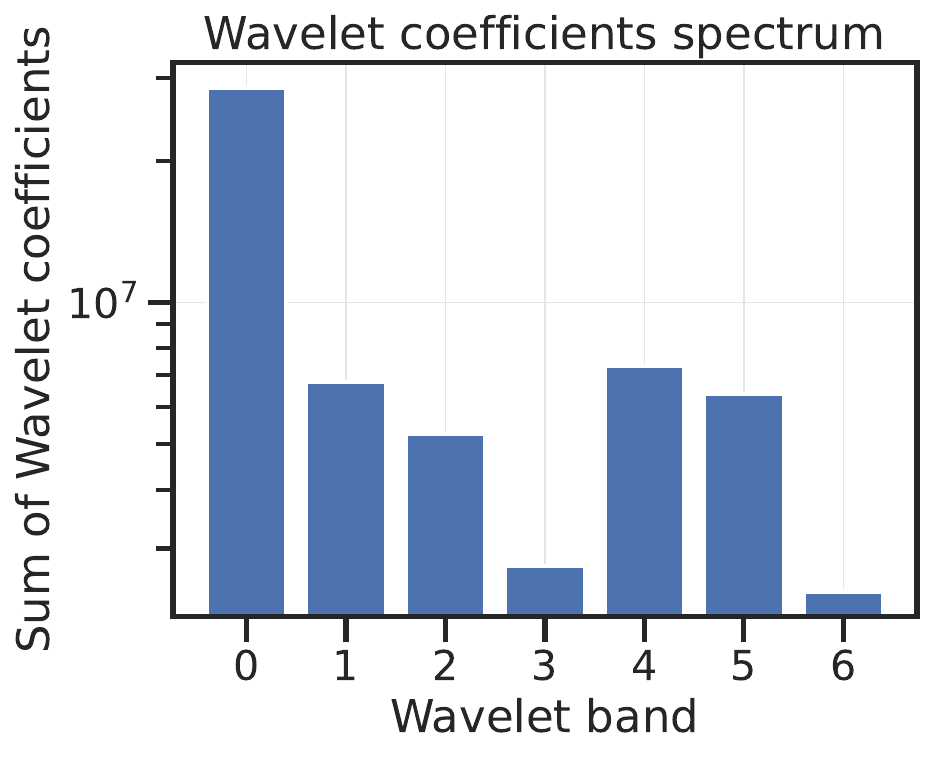}
    \caption{Sum of singular values, Fourier and wavelet coefficients in different bands.}
    \label{fig:explained_variance_power}
\end{figure}

\subsection{Decay in power or spectral density}

We sort the eigenvalues, power and wavelet coefficients in decreasing order and plot them. We find that they decay exponentially for all 3 linear bases.

\begin{figure}[H]
    \centering
    \includegraphics[width=0.32\textwidth]{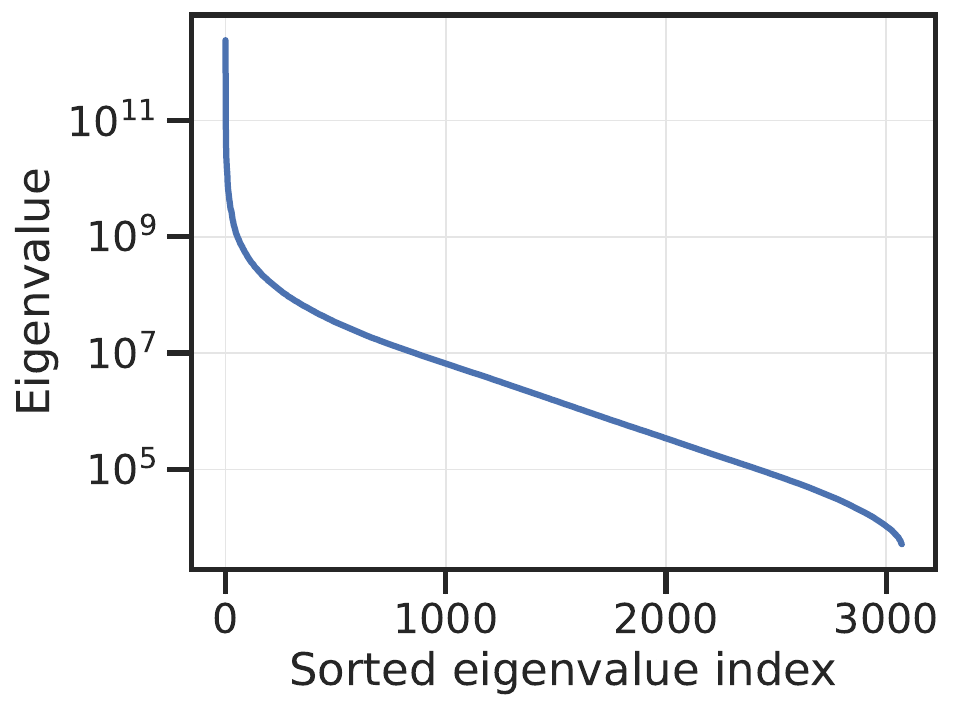}
    \includegraphics[width=0.32\textwidth]{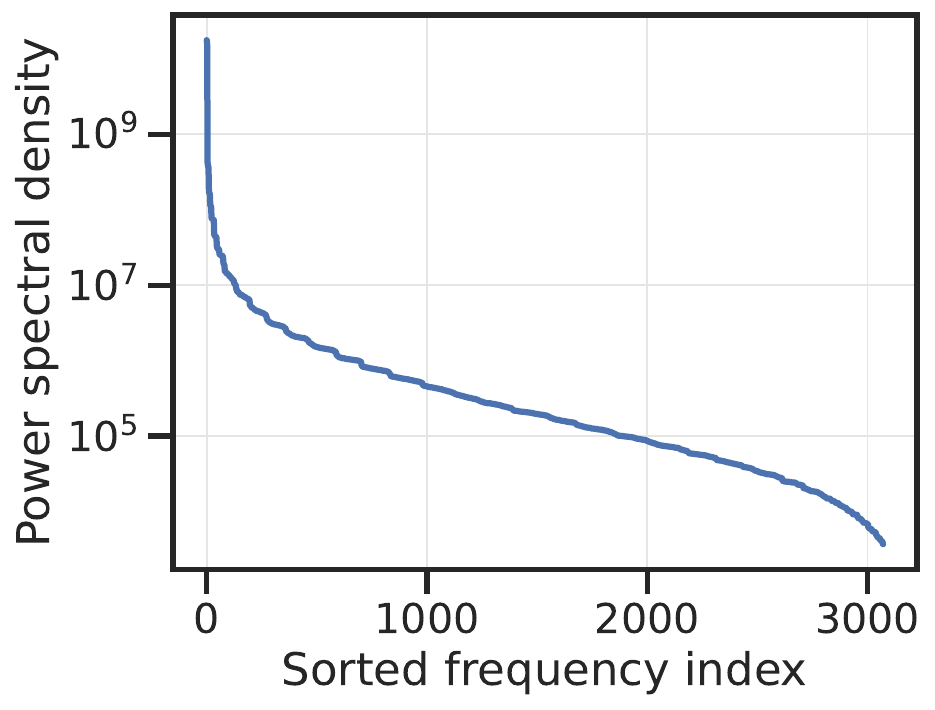}
    \includegraphics[width=0.32\textwidth]{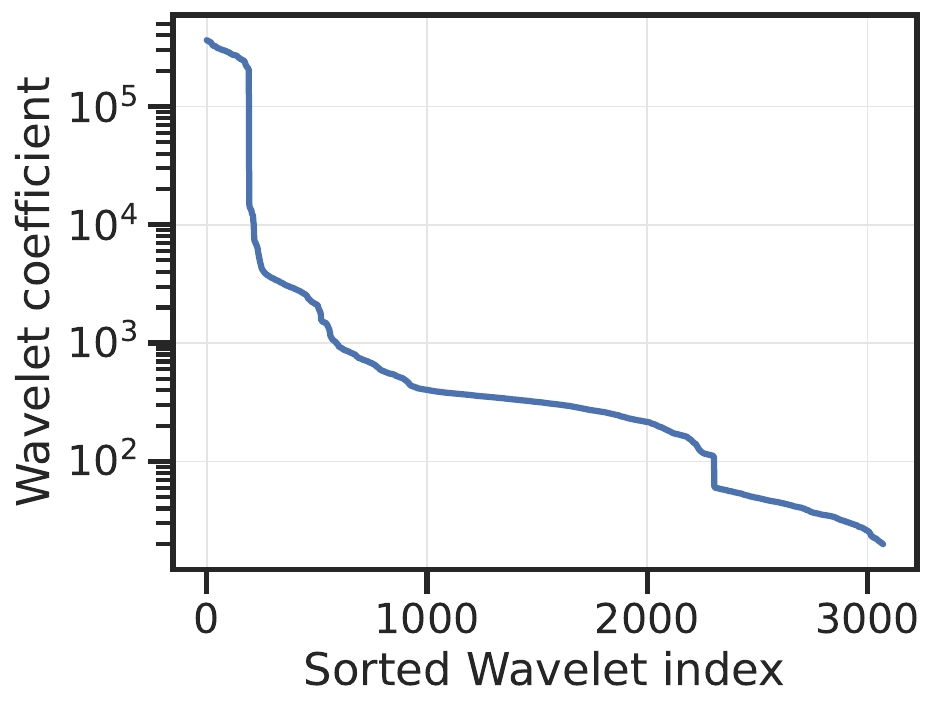}
    \caption{The decay in coefficients of the Eigenvalues frequencies and wavelets. The spectra of all 3 linear bases decay exponentially and have a characteristic small head and a long tail.}
    \label{fig:spectral_decay}
\end{figure}

\subsection{Additional Experiments on Trained Network Frequency Sensitivity}

\begin{figure}[H]
    \centering
    \includegraphics[width=0.325\linewidth]{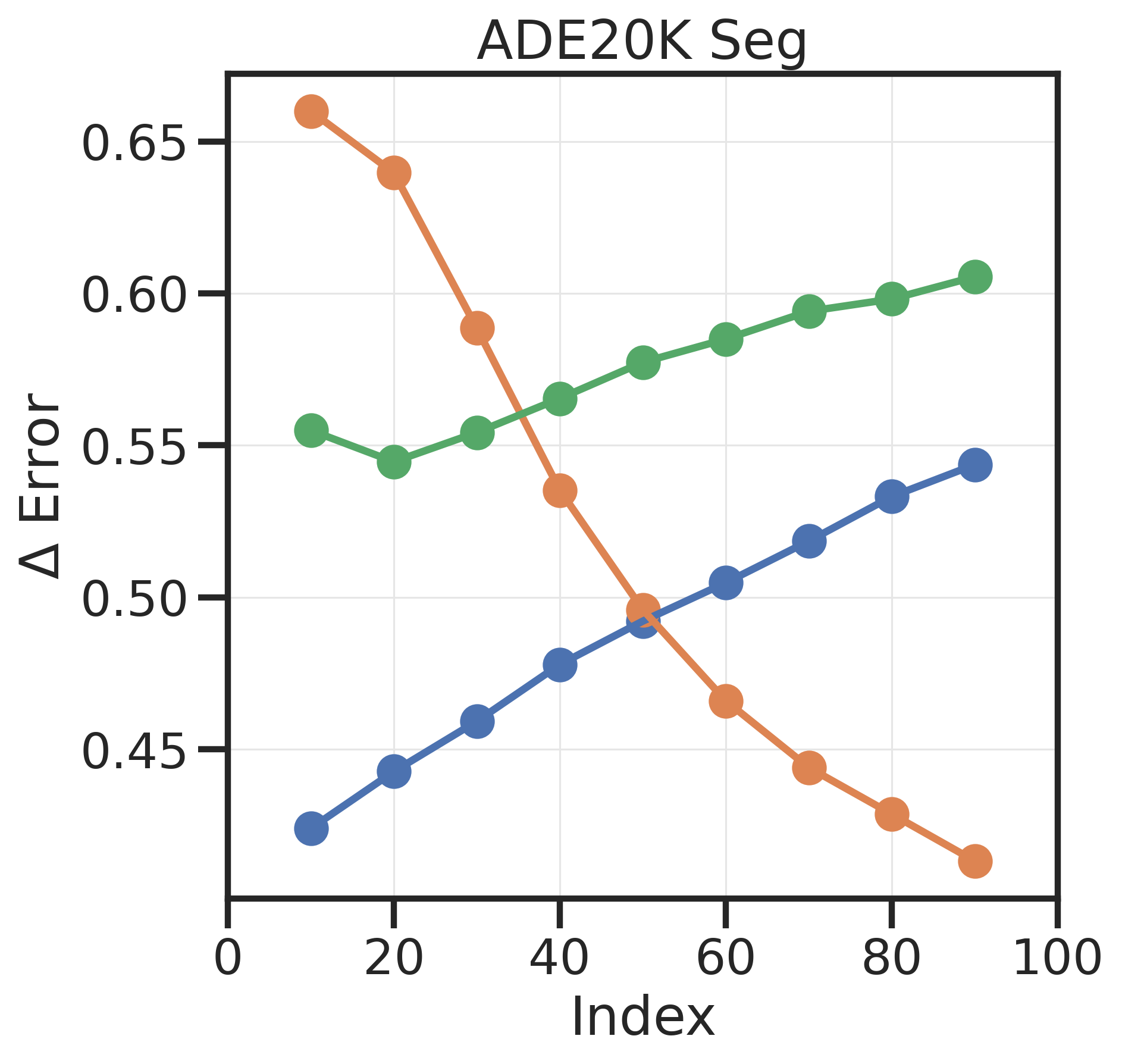}
    \includegraphics[width=0.325\linewidth]{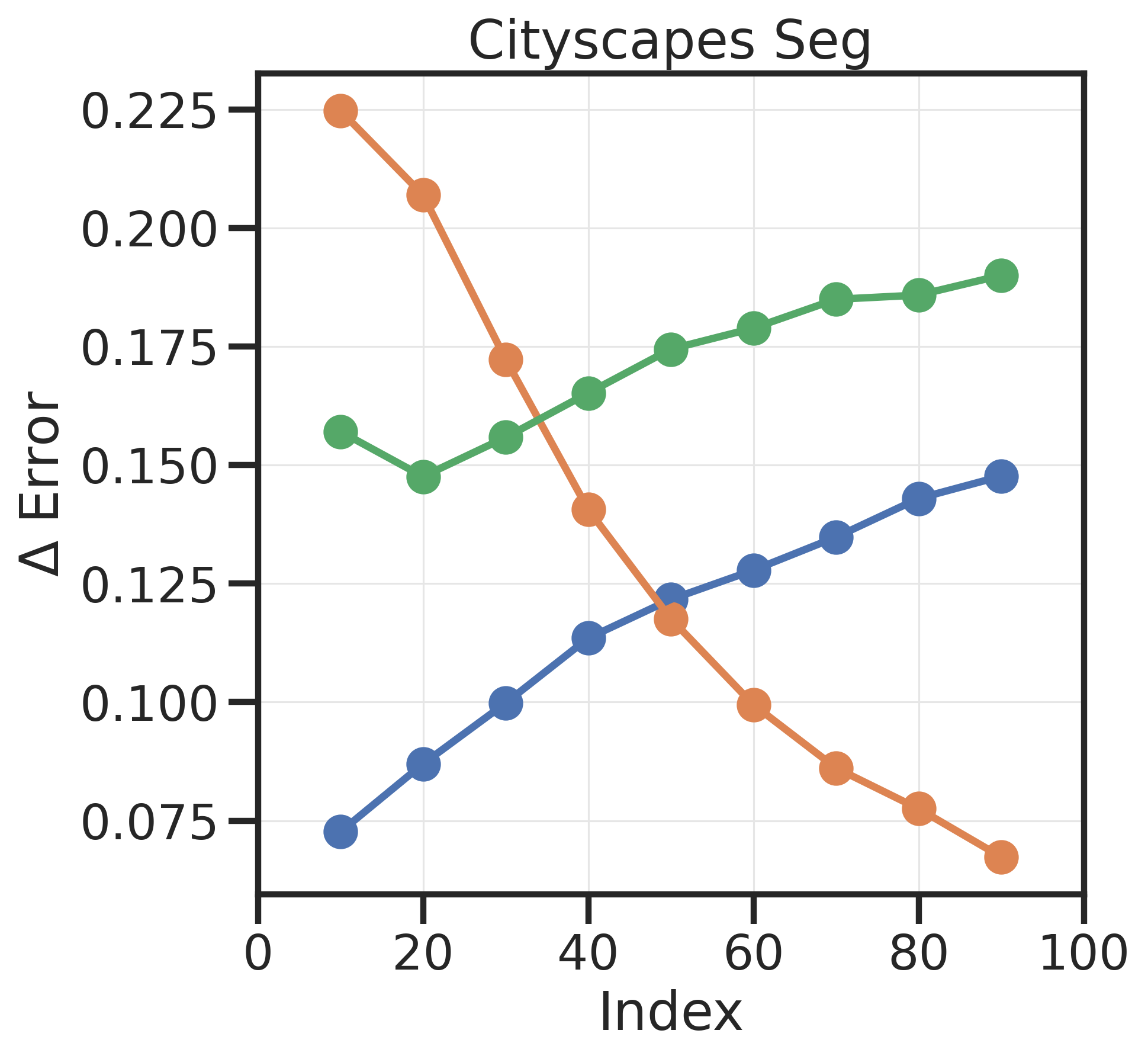}
    \includegraphics[width=0.325\linewidth]{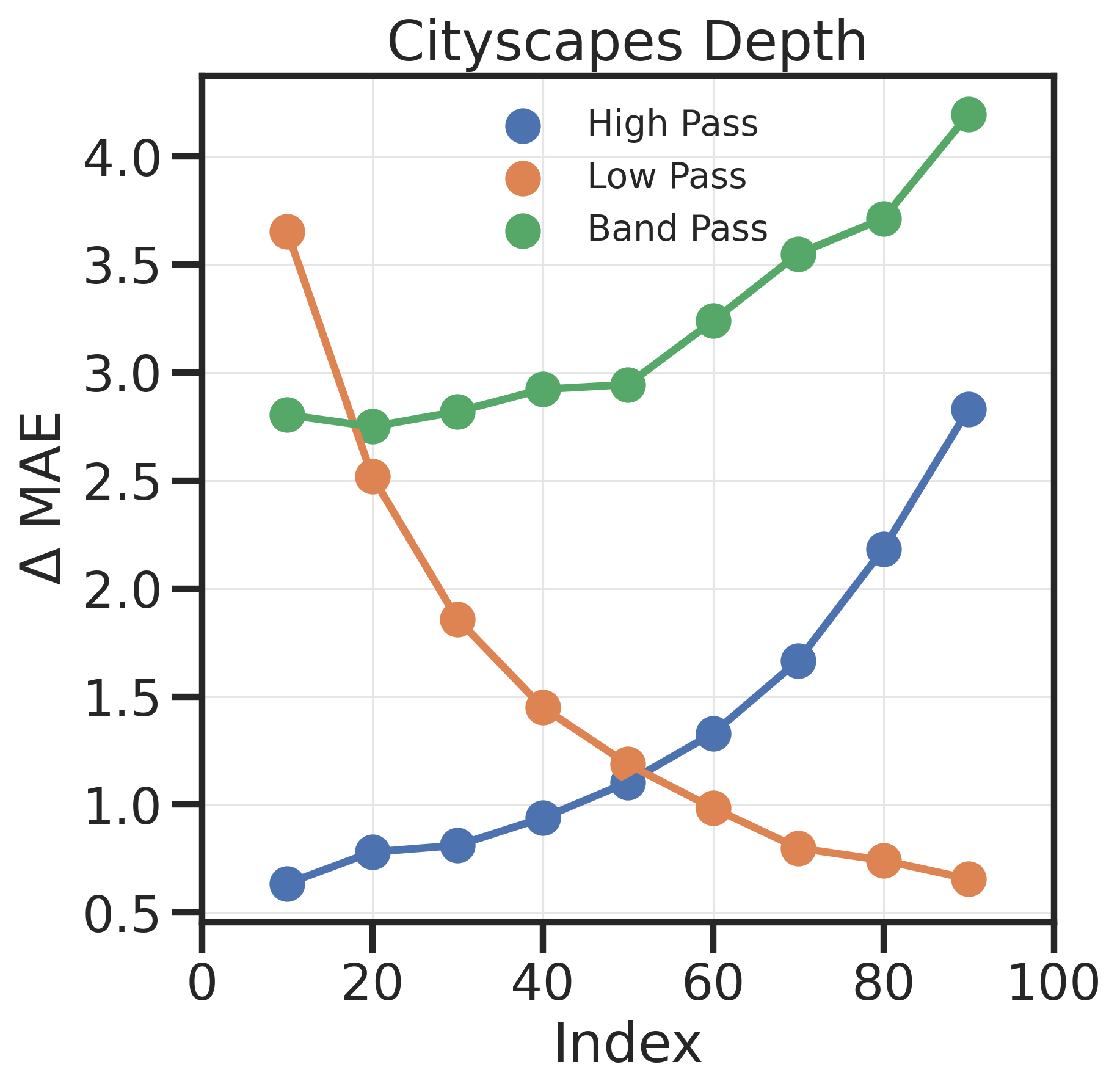}
    \caption{We showcase some additional experiments of taking a pretrained network and applying various filters for inference to the network and observing network performance. We find a similar analysis to other experiments in the paper showcasing a bias towards using components of the low-frequency prediction. }
    \label{fig:sensitvity_extra}
\end{figure}

\section{Additional Details for Methods}

\subsection{Relationship between PCA and the Fourier basis}
\label{s:PCA}
We provide the proof from \cite{natural_image_stats} that shows the eigenvectors of the pixel-wise covariance matrix correspond to sinusoids with phase $\alpha$. Assume that the entries of the covariance matrix corresponding to two pixels $x, x'$ only depend on the distance between them. This proof can be generalized to different sinusoid frequencies. Hence, sinusoids are eigenvectors of the covariance matrix and are ordered based on their natural distribution in the data. 
\begin{equation*}
\begin{aligned}
\sum_{x} \text{cov}(x, x') \sin(x + \alpha) &= \sum_{x} c(x - x') \sin(x + \alpha) = \sum_{z} c(z) \sin(z + x' + \alpha) \\*
\sum_{z} c(z) \sin(z + x' + \alpha) &= \sum_{z} c(z) (\sin(z) \cos(x' + \alpha) + \cos(z) \sin(x' + \alpha)) \\*
&= \left[ \sum_{z} c(z) \sin(z) \right] \cos(x' + \alpha) + \left[ \sum_{z} c(z) \cos(z) \right] \sin(x' + \alpha) \\*
\sum_{x} \text{cov}(x, x') \sin(x + \alpha) &= \left[ \sum_{z} c(z) \cos(z) \right] \sin(x' + \alpha)
\end{aligned}
\end{equation*}

\end{document}